\documentclass[11pt]{article}

\usepackage[preprint]{acl}
\usepackage{amsmath} 
\usepackage{times}
\usepackage{latexsym}
\usepackage{amssymb}

\usepackage[T1]{fontenc}
\usepackage[utf8]{inputenc}

\usepackage{microtype}

\usepackage{inconsolata}
\usepackage{caption}
\usepackage{cuted}
\usepackage{xspace}

\usepackage{graphicx}
\usepackage{xcolor}
\usepackage{booktabs}
\usepackage{multirow}
\usepackage{tabularx}
\usepackage{array}
\usepackage{makecell}
\usepackage{subcaption}
\usepackage{colortbl}
\usepackage{pgfplots}
\pgfplotsset{compat=1.18}
\definecolor{trainbg}{HTML}{FFF3CD}  % highlight colour for training config cell
\usepackage{tcolorbox}
\tcbuselibrary{breakable} % <-- Load the breaking library

\newtcolorbox{promptbox}[1][]{
    breakable,             % <-- Tells the box it is allowed to split across pages
    colback=gray!4!white,
    colframe=gray!60!black,
    fonttitle=\bfseries,
    coltitle=white,
    colbacktitle=gray!70!black,
    boxrule=0.5pt,
    arc=3pt,
    left=8pt,
    right=8pt,
    top=6pt,
    bottom=6pt,
    boxsep=0pt,
    toptitle=3pt,
    bottomtitle=3pt,
    title=#1
}

\newcommand{\bad}[1]{\cellcolor{red!10}\textcolor{red!60!black}{#1}}
\newcommand{\warn}[1]{\cellcolor{yellow!20}\textbf{#1}}
\newcommand{\good}[1]{\cellcolor{green!15}\textbf{#1}}
\newcommand{\methodname}{\textsc{Frames2LoRA}\xspace}
\newcommand{\adjustimg}{\hspace*{\dimexpr\evensidemargin-\oddsidemargin}}
\newcommand{\centerimg}[2][width=\textwidth]{\makebox[\textwidth]{\adjustimg\includegraphics[#1]{#2}}}

\title{\methodname: Parametric Video Internalization \\ for Vision-Language Models}

\author{
  Manan Suri$^{\dagger}$\thanks{Equal contribution.},
  Sarvesh Baskar\footnotemark[1],
  Dinesh Manocha$^{\dagger}$ \\
  $^{\dagger}$University of Maryland, College Park \\
  \texttt{manans@umd.edu} \quad \texttt{baskarsarvesh@gmail.com} \\ \\
  \url{https://frames2lora.github.io/}
}

\begin{document}

\maketitle

\begin{strip}
  \noindent\centerimg[width=\linewidth]{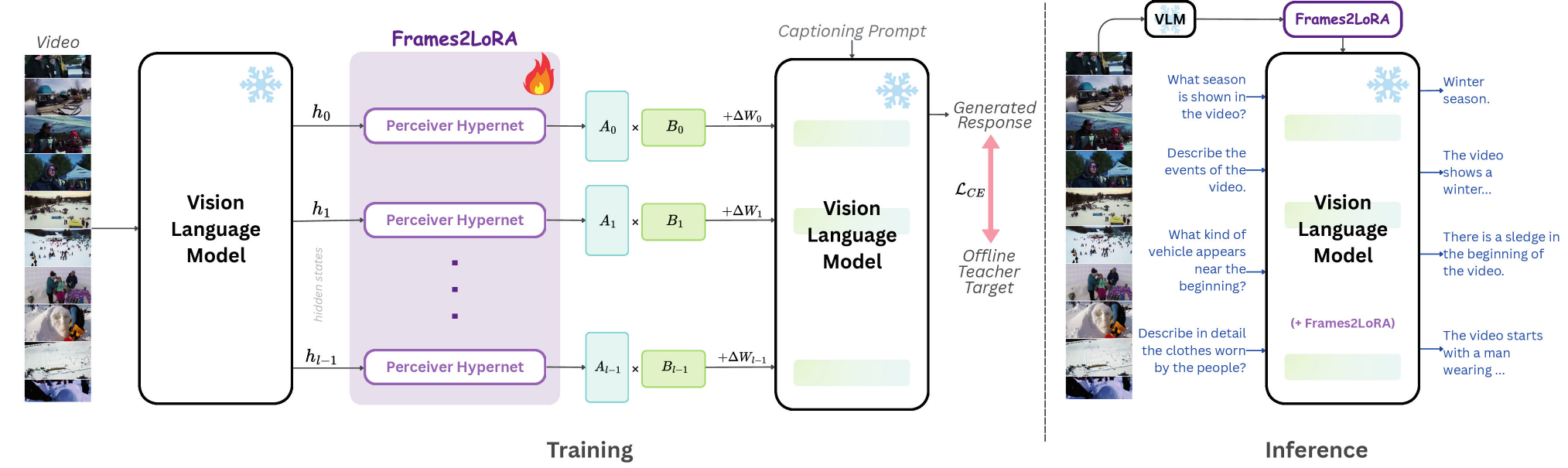}
  \fontsize{10pt}{12pt}\selectfont
  Figure 1: \textbf{\methodname \space overview.}
  \textit{Training (left):} A frozen VLM encodes the input video into hidden states.
  The trainable \methodname hypernetwork reads these states and generates LoRA adapter weights in a single forward pass.
  The adapter-augmented frozen VLM is trained against teacher-generated targets.
  \textit{Inference (right):} Given a new video, \methodname generates the LoRA adapter once.
  The frozen VLM, augmented with this adapter, answers arbitrary text queries without visual tokens.
  Per-query cost is independent of video length.
  \label{fig:overview}
\end{strip}
\setcounter{figure}{1}

% ---------------------------------------------------------------
\begin{abstract}
% ---------------------------------------------------------------
Processing video in vision-language models is expensive: each frame
occupies hundreds of tokens, and inference cost scales with every frame
and every repeated query.
We introduce \textbf{\methodname}, a method for parametric video
internalization.
A perceiver hypernetwork reads the intermediate representations produced
layer-by-layer as a frozen VLM encodes a video, and generates a
Low-Rank Adaptation (LoRA) adapter in a single forward pass.
Unlike standard LoRA fine-tuning, which requires iterative gradient
updates, \methodname predicts these weights directly from the video.
Trained for SmolVLM2 500M and 2.2B on video summarization and captioning,
\methodname enables the same frozen VLM to answer queries from the adapter
alone, with \textit{\textbf{zero visual tokens}} in its context at query time.
\methodname is statistically non-inferior and equivalent to direct
video-in-context inference across all five captioning benchmarks at both
model scales, and across seven of eight video question answering
benchmark-scale pairings.
Although trained only on 12 frames at 384px, it remains stable up to
1{,}024 frames and 1024px, where direct video-in-context inference often
degenerates. Across this sweep, it reduces answer-time visual-token load by up
to 1{,}500$\times$ and query TTFT by 6--80$\times$, while preserving
video-faithful outputs. We also find that independently generated adapters for
non-overlapping video segments can compose in rank space, suggesting a path
toward chunked long-video internalization.
\end{abstract}

\section{Introduction}
% ---------------------------------------------------------------

Video understanding in VLMs is built on a token-heavy abstraction:
frames are encoded as visual tokens and concatenated into the model's
context window.
Each frame at standard resolution contributes hundreds of visual
tokens~\citep{liu2024llava15,shang2024llavaprumerge}; even short clips of
a few dozen frames generate tens of thousands of tokens before any text
query is added, and memory and latency scale with every frame and every
query.
Past a capacity threshold, this bottleneck does not produce gracefully
degraded outputs: VLMs generate incoherent or repetitive text unrelated
to the video~\citep{xue2024longvila,zhang2024longva}.
The context window (the model's fixed token capacity) is therefore the
fundamental bottleneck for video understanding, and it is re-encountered
on every query over the same video.

Much work aims to fit more video into the context window.
Frame subsampling~\citep{zhang2023videollama} discards frames to meet a
token budget, sacrificing temporal coverage.
Visual token compression methods~\citep{shang2024llavaprumerge,
li2024tokenpacker} prune or merge spatial tokens before the language
backbone, reducing per-frame cost without discarding entire frames.
Long-context architectures~\citep{xue2024longvila,zhang2024longva} scale
the context window itself through sequence parallelism and position
encoding modifications.
Streaming methods~\citep{qian2024streaming} process video incrementally,
maintaining a compact memory buffer in lieu of full context retention.
Each approach reduces the burden without resolving it: visual tokens remain
in context at query time, every query re-incurs the encoding overhead, and
all approaches eventually encounter the same capacity ceiling.
The capacity ceiling is not a constraint to manage: it is a constraint to
eliminate.

We take a fundamentally different approach.
Rather than compressing visual information to fit within the context window,
we eliminate it from the query entirely, encoding the video into the model's
parameters before any query is issued.
The video is stored as a LoRA adapter~\citep{hu2022lora}; subsequent
queries are answered by a frozen base model with those adapter
weights, with no visual tokens in context.
Prior work has shown that feedforward
hypernetworks~\citep{ha2016hypernetworks,charakorn2026doc2lora} can produce
LoRA adapters from \emph{text documents}, enabling a frozen LLM to answer
queries about a document with no text tokens in context.
Extending this paradigm to video introduces qualitatively harder
challenges: the token volume per example is orders of magnitude larger,
making iterative per-example optimization computationally impractical; the
compression is cross-modal, requiring visual semantics to be expressed as
perturbations to a language model's parameter space; and the visual
distribution varies along a resolution axis with no textual analog.

\noindent {\bf Main Result:} We introduce \textbf{\methodname}, a framework for parametrically internalizing videos into a frozen vision-language model (VLM). Given a video, a perceiver hypernetwork~\citep{jaegle2021perceiverio} processes the layer-wise hidden states of the frozen VLM encoder and generates LoRA adapter weights in a single forward pass. The generated adapter is then attached to the same frozen VLM, enabling it to answer questions about the video without requiring visual tokens in the context window. During training, both the VLM encoder and the answering model remain frozen; only the hypernetwork is optimized using cached teacher-generated captions and summaries as supervision. We train and evaluate \methodname on SmolVLM2 500M and 2.2B~\citep{marafioti2025smolvlm}.
Our novel contributions include:
\begin{itemize}

\item \textbf{First parametric video internalization.}
A Perceiver hypernetwork that converts a video into a LoRA adapter in a
single forward pass, enabling a frozen VLM to answer queries with no visual
tokens in context.
We demonstrate feasibility across  2.2B and 500M model scales.

\item \textbf{Strong performance on captioning and video question answering.}
Statistical non-inferiority and equivalence to direct video-in-context
inference across all five captioning benchmarks at both model scales
(ActivityNet~Captions, PLM-RDCap, PLM-RCap, VDC, CaReBench) and across
seven of eight video question answering benchmark-scale pairings
(NExT-QA, ActivityNet-QA, PLM-SGQA, VidCapBench).

\item \textbf{Efficiency, generalization and emergent compositionality.}
Although trained only on 12 frames at 384px, \methodname remains stable
up to 1{,}024 frames and 1024px, where direct video-in-context inference often
degenerates. It reduces answer-time visual-token load by up to
1{,}500$\times$ and query TTFT by 6--80$\times$, while preserving video-faithful
outputs. Compared to KV caching and token-compression techniques, we show that video internalization via \methodname preserves performance across token budgets, is faster to process, and has the lowest time to first token. We further observe that adapters generated independently for
non-overlapping video segments can compose in rank space, suggesting a path
toward chunked long-video internalization. 

\end{itemize}

\section{Related Work}
% ---------------------------------------------------------------

\subsection{Efficient Video Understanding}

Most efficient video-understanding methods reduce the number or cost of visual
tokens while still keeping visual information in the model context. Frame
subsampling~\citep{zhang2023videollama} lowers temporal coverage to fit a token
budget; visual-token compression~\citep{shang2024llavaprumerge,
li2024tokenpacker} prunes or merges spatial tokens; long-context video
models~\citep{xue2024longvila,zhang2024longva} extend the usable context
window; and streaming methods~\citep{qian2024streaming,zhang2023videollama}
maintain compact memory across time. These approaches improve scalability, but
the language model still conditions on visual tokens at query time.
\methodname is orthogonal: it converts the video into adapter weights
once, then answers later queries without visual tokens in context.

\subsection{Parametric Knowledge Compression}

Parameter-efficient methods such as LoRA, prefix tuning, and prompt tuning store
task information in small learned updates rather than full model
parameters~\citep{hu2022lora,li2021prefixtuning,lester2021prompttuning}. More
recent work moves instance-level context into compact representations, including
gist tokens~\citep{mu2023gisting}, hypernetwork-based editing
\citep{mitchell2021mend,ha2016hypernetworks}, and deep context distillation
\citep{caccia2025deepcd}. Closest to our setting, Doc-to-LoRA maps text
documents into LoRA adapters using a feedforward hypernetwork
\citep{charakorn2026doc2lora}. \methodname extends this idea from text
to video, where the hypernetwork must compress high-volume visual context into
language-model adapter weights and generalize across frame count and resolution.

\section{\methodname}
\label{sec:frames2lora}

\methodname converts a video into a video-specific LoRA adapter in a
single forward pass. A frozen VLM encodes the video into layer-wise hidden
states, and a trainable Perceiver hypernetwork maps these states into LoRA
weights. At inference time, the generated adapter is attached to the frozen
answer model, which answers downstream text prompts without receiving any
visual tokens in its context.

\subsection{Problem Formulation}

Let $v$ denote a video, $i$ an internalization instruction, $p$ a downstream
text prompt, and $y$ the target response. We assume a frozen vision-language
encoder $E$, a frozen answer model $F$, and a trainable hypernetwork
$H_\phi$. The method is defined as:
\begin{align}
    \mathbf{C} &= E(v, i), \label{eq:v2l_encode} \\
    \theta(v) &= H_\phi(\mathbf{C}), \label{eq:v2l_generate} \\
    p_\phi(y \mid p, v) &= F(y \mid p;\theta(v)). \label{eq:v2l_answer}
\end{align}
Here, $\mathbf{C}$ denotes video-conditioned hidden states and $\theta(v)$
denotes the generated LoRA adapter. The answer model receives the text prompt
$p$ and the adapter $\theta(v)$, but not the video tokens. During training,
only $\phi$ is updated; both $E$ and $F$ remain frozen.

\subsection{Video Encoder}

We use a frozen SmolVLM2 model~\citep{marafioti2025smolvlm} as the video
encoder. Given a sampled video and the internalization instruction, we collect
the text-side hidden states from each transformer layer:
\begin{equation}
    \mathbf{C}
    =
    \mathrm{stack}(\mathbf{h}_0,\mathbf{h}_1,\ldots,\mathbf{h}_{L-1})
    \in \mathbb{R}^{L \times S \times D},
    \label{eq:v2l_context}
\end{equation}
where $L$ is the number of layers, $S$ is the fused sequence length, and $D$ is
the hidden dimension. Keeping the layer dimension allows the hypernetwork to
generate layer-indexed adapters instead of using a single pooled video vector
for all layers.

\subsection{Perceiver Hypernetwork}

The hypernetwork maps $\mathbf{C}$ into LoRA weights for selected linear
modules of the frozen model. We use a Perceiver-style resampler
architecture~\citep{jaegle2021perceiverio}. For each layer slice
$\mathbf{C}_\ell \in \mathbb{R}^{S \times D}$, an encoder resampler attends
from learned latent queries to the video-conditioned hidden states, producing a
fixed-size representation. A decoder resampler then uses one output query for
each target module and LoRA rank direction.

For batch size $B$, number of target modules $M$, rank $R$, and latent size
$Z$, the hypernetwork output has shape
\begin{equation}
    \mathbf{O}
    \in
    \mathbb{R}^{B \times L \times M \times R \times Z}.
    \label{eq:v2l_hyper_out}
\end{equation}
A shared projection head maps each rank latent to the two LoRA factors:
% \begin{align}
%     \mathbf{A}_{\ell,m}
%     &\in \mathbb{R}^{R \times d_{\mathrm{in}}}, \\
%     \mathbf{B}_{\ell,m}
%     &\in \mathbb{R}^{R \times d_{\mathrm{out}}},
% \end{align}
\begin{equation}
\begin{aligned}
    \mathbf{A}_{\ell,m} &\in \mathbb{R}^{R \times d_{\mathrm{in}}}, \\
    \mathbf{B}_{\ell,m} &\in \mathbb{R}^{R \times d_{\mathrm{out}}}.
\end{aligned}
\end{equation}
where $\ell$ indexes the transformer layer and $m$ indexes the target linear
module. The generated factors are scaled by learned multipliers, with the
$\mathbf{A}$ scale initialized to one and the $\mathbf{B}$ scale initialized to
zero. 

\subsection{Dynamic LoRA Injection}

For a frozen linear layer with weight
$\mathbf{W} \in \mathbb{R}^{d_{\mathrm{out}} \times d_{\mathrm{in}}}$, we use
the standard LoRA factorization~\citep{hu2022lora}. Under the row-vector
implementation convention, the frozen layer computes
$\mathbf{x}\mathbf{W}^{\top}$. The generated adapter adds:
\begin{equation}
    \Delta \mathbf{y}
    =
    s \, (\mathbf{x}\mathbf{A}_{\ell,m}^{\top})\mathbf{B}_{\ell,m},
    \label{eq:v2l_lora_delta}
\end{equation}
where $s$ is the fixed LoRA scaling factor. The full adapted forward pass is:
\begin{equation}
    \mathbf{y}
    =
    \mathbf{x}\mathbf{W}^{\top}
    +
    s \, (\mathbf{x}\mathbf{A}_{\ell,m}^{\top})\mathbf{B}_{\ell,m}.
    \label{eq:v2l_lora_forward}
\end{equation}
Each example receives its own generated adapter, so the LoRA weights are
conditioned on the input video rather than shared across all videos.

\subsection{Training Objective}

We train the hypernetwork with teacher-forced cross-entropy over response
tokens:
\begin{equation}
    \mathcal{L}(\phi)
    =
    -\sum_t
    \log p_\phi(y_t \mid y_{<t}, p, \theta(v)).
    \label{eq:v2l_loss}
\end{equation}
The answer model receives only the downstream text prompt and the generated
adapter during this loss computation. 

% \subsection{Training Targets}

% We train on video spans derived from FineVideo~\citep{Farre2024FineVideo}.
% The dataset metadata is used to define temporal spans, including single-scene
% spans, adjacent multi-scene spans, and full-video spans. It is not used as the
% final supervision target. Instead, for each span, we generate a visual-only
% teacher response offline using a frozen SmolVLM2 teacher conditioned on the
% sampled video frames and the downstream prompt. These cached teacher responses
% serve as the targets for cross-entropy training. Audio is excluded throughout.

% ---------------------------------------------------------------

\section{Experimental Setup}
\label{sec:experimental_setup}

\subsection{Models and Training}

We evaluate two SmolVLM2 model scales: 500M and 2.2B. For each scale, the
video encoder and answer model are initialized from the same frozen backbone.
Only the \methodname hypernetwork is trained. Training uses 12 uniformly sampled frames at 384px longest-edge resolution (constrained by compute). We
apply generated LoRA adapters to the MLP \texttt{down\_proj} modules of the text decoder, with rank
$R=16$. We train on video spans derived from FineVideo~\citep{Farre2024FineVideo}.
The span mixture contains single-scene spans, adjacent multi-scene spans, and
full-video spans, sampled in a 60/30/10 ratio. FineVideo metadata is used only
to define spans; the final training targets are cached offline teacher
generations from a frozen SmolVLM2 teacher conditioned on the sampled video
frames and downstream prompt. Audio is excluded throughout. The hypernetwork is
trained with teacher-forced cross-entropy over response tokens, while the
answer model receives only the text prompt and generated adapter. Further details on training can be found in the appendix.

% \subsection{Compared Systems}

% The primary baseline is direct video-in-context inference with the frozen
% SmolVLM2 model. For each example, the baseline receives the sampled video frames
% and the downstream text prompt in context. \methodname first converts
% the video into a LoRA adapter using the frozen encoder and then answers the same
% prompt using only text tokens and the generated adapter. Thus, both systems use
% the same frozen backbone and prompts; they differ only in whether video
% information is supplied as query-time context tokens or as generated adapter
% weights.

\begin{table*}[ht]
\centering
\resizebox{0.8\textwidth}{!}{%
\renewcommand{\arraystretch}{1.12}
\setlength{\tabcolsep}{3.0pt}
\begin{tabular}{l  r r r r c c  r r r r c c}
\toprule
\multicolumn{13}{l}{\textbf{LLM Judge}} \\
\cmidrule(r){1-13}
\multirow{2}{*}{\textbf{Benchmark}}
  & \multicolumn{6}{c}{\textbf{SmolVLM 500M}}
  & \multicolumn{6}{c}{\textbf{SmolVLM 2.2B}} \\
\cmidrule(lr){2-7}\cmidrule(lr){8-13}
& Base & V2L & $\Delta$ & CI & Eq & NI & Base & V2L & $\Delta$ & CI & Eq & NI \\
\midrule
ActivityNet~Captions & 0.428 & 0.356 & -0.072 & [-0.104, -0.041] & \good{Y} & \good{Y} & 0.576 & 0.492 & -0.084 & [-0.113, -0.057] & \good{Y} & \good{Y} \\
PLM-RDCap & 0.308 & 0.263 & -0.045 & [-0.069, -0.021] & \good{Y} & \good{Y} & 0.326 & 0.316 & -0.010 & [-0.032, +0.012] & \good{Y} & \good{Y} \\
PLM-RCap & 0.252 & 0.242 & -0.011 & [-0.031, +0.009] & \good{Y} & \good{Y} & 0.270 & 0.287 & +0.017 & [+0.001, +0.034] & \good{Y} & \good{Y} \\
VDC (aggregate) & 0.515 & 0.406 & -0.108 & [-0.118, -0.098] & \good{Y} & \good{Y} & 0.539 & 0.511 & -0.028 & [-0.037, -0.019] & \good{Y} & \good{Y} \\
CaReBench & 0.334 & 0.278 & -0.056 & [-0.067, -0.045] & \good{Y} & \good{Y} & 0.437 & 0.369 & -0.068 & [-0.078, -0.058] & \good{Y} & \good{Y} \\
\midrule
\textit{Average} & 0.367 & 0.309 & -0.058 & [-0.078, -0.039] & \good{Y} & \good{Y} & 0.430 & 0.395 & -0.035 & [-0.052, -0.018] & \good{Y} & \good{Y} \\
\midrule[0.8pt]
\multicolumn{13}{l}{\textbf{Token F1}} \\
\cmidrule(r){1-13}
\multirow{2}{*}{\textbf{Benchmark}}
  & \multicolumn{6}{c}{\textbf{SmolVLM 500M}}
  & \multicolumn{6}{c}{\textbf{SmolVLM 2.2B}} \\
\cmidrule(lr){2-7}\cmidrule(lr){8-13}
& Base & V2L & $\Delta$ & CI & Eq & NI & Base & V2L & $\Delta$ & CI & Eq & NI \\
\midrule
ActivityNet~Captions & 0.236 & 0.243 & +0.007 & [+0.002, +0.012] & \good{Y} & \good{Y} & 0.263 & 0.256 & -0.007 & [-0.012, -0.002] & \good{Y} & \good{Y} \\
PLM-RDCap & 0.189 & 0.198 & +0.009 & [+0.005, +0.013] & \good{Y} & \good{Y} & 0.198 & 0.207 & +0.009 & [+0.005, +0.013] & \good{Y} & \good{Y} \\
PLM-RCap & 0.177 & 0.203 & +0.026 & [+0.021, +0.031] & \good{Y} & \good{Y} & 0.199 & 0.204 & +0.005 & [+0.001, +0.010] & \good{Y} & \good{Y} \\
VDC (aggregate) & 0.315 & 0.288 & -0.027 & [-0.030, -0.025] & \good{Y} & \good{Y} & 0.297 & 0.304 & +0.007 & [+0.003, +0.010] & \good{Y} & \good{Y} \\
CaReBench & 0.295 & 0.275 & -0.020 & [-0.023, -0.017] & \good{Y} & \good{Y} & 0.292 & 0.279 & -0.013 & [-0.015, -0.010] & \good{Y} & \good{Y} \\
\midrule
\textit{Average} & 0.243 & 0.242 & -0.001 & [-0.005, +0.003] & \good{Y} & \good{Y} & 0.250 & 0.250 & +0.000 & [-0.004, +0.004] & \good{Y} & \good{Y} \\
\bottomrule
\end{tabular}
}
\caption{
Comparison of the base model with video and \methodname generated adapters, across captioning benchmarks using LLM Judge scores and Token F1. We report mean scores, the paired difference $\Delta$ (V2L $-$ Base), 95\% confidence intervals, and the statistical equivalence (Eq) and non-inferiority (NI) criteria.
}
\label{tab:cap_main}
\end{table*}

% \subsection{Evaluation Benchmarks}

% We evaluate captioning on ActivityNet Captions~\citep{krishna2017activitynet},
% PLM-RDCap \cite{cho2025PerceptionLM}, PLM-RCap \cite{cho2025PerceptionLM}, the Video Description Corpus (VDC) \cite{chai2024auroracap}, and CaReBench \cite{xu2024carebench}. 
% % \\\\
% We evaluate video question answering on NExT-QA \cite{xiao2021nextqa}, ActivityNet-QA \cite{yu2019activitynetqa}, PLM-SGQA \cite{cho2025PerceptionLM}, and
% VidCapBench \cite{chen2025vidcapbench}. For all benchmarks, the direct baseline and \methodname
% use the same videos, prompts, references, frame sampling, and decoding
% configuration.

\subsection{Evaluation Benchmarks}
\label{sec:eval_benchmarks}

We evaluate captioning on ActivityNet Captions~\citep{krishna2017activitynet},
PLM-RDCap~\citep{cho2025PerceptionLM}, PLM-RCap~\citep{cho2025PerceptionLM},
VDC~\citep{chai2024auroracap}, and CaReBench~\citep{xu2024carebench}; and
video QA on NExT-QA~\citep{xiao2021nextqa}, ActivityNet-QA~\citep{yu2019activitynetqa},
PLM-SGQA~\citep{cho2025PerceptionLM}, and
VidCapBench~\citep{chen2025vidcapbench}.

To scale LLM Judge evaluation, we fix the number of samples from each benchmark to 500. VDC and CaReBench use
500 examples per subset/style. VidCapBench has multiple QA pairs corresponding to each video, therefore we fixed the number of videos to 100 and obtained 1{,}523 QA pairs corresponding to it. For all benchmarks, the direct baseline and
\methodname use the same videos, prompts, references, frame sampling,
and decoding configuration.

\subsection{Metrics and Statistical Testing}

We report two quality metrics. First, we compute token-level F1 between the
generated output and the reference answer or caption. Second, we use an LLM
judge to score output quality on a 1--5 scale, which is linearly rescaled to
$[0,1]$. We use Qwen3-30B \cite{qwen3} as our judge model, with a constrained rubric. Human study on this metric for a subset of 200 examples (100 captioning + 100 QA) reveals strong correlation with human judgements, with Spearman $\rho = 0.823$ for metric fidelity. 
% \\\\
% We also report output preservation. In this setting, the judge compares
% \methodname's output directly against the direct video-in-context
% baseline output for the same video and prompt, without using the dataset
% reference. This measures whether the generated adapter preserves the frozen
% model's own video-conditioned behavior.
% \\\\

We estimate 95\% confidence intervals using paired bootstrap resampling.
For statistical measures, \textit{NI} (Non-inferiority) and \textit{Eq} (Equivalence) we use a margin of $0.05$ for
token-F1 and $0.15$ for rescaled judge score.

% \subsection{Generalization, Efficiency, and Composition}

% To test generalization beyond the training setting, we evaluate the trained
% hypernetwork without further updates across frame counts
% $\{8,12,24,48,128,256,512,1024\}$ and longest-edge resolutions
% $\{224,336,512,1024\}$. We report paired quality deltas, query-time speedups,
% and input-token reductions relative to direct video-in-context inference.
% \\\\
% For efficiency, we measure time-to-first-token (TTFT). Direct inference
% re-encodes the video for every query. \methodname separates the one-time
% internalization step from text-only answering, so the generated adapter can be
% reused across repeated questions about the same video. We therefore report both
% single-query TTFT and amortized per-query TTFT as the number of questions per
% video increases.
% \\\\
% Finally, we evaluate chunk composition by splitting each video into two
% temporal halves, generating one adapter per half, and concatenating the adapters
% along the LoRA rank dimension before generation. This tests whether
% independently internalized segments can be combined without additional
% composition-specific training.

% ---------------------------------------------------------------
% ---------------------------------------------------------------

\section{Results}
\label{sec:results}

% We evaluate \methodname along three main axes: output quality,
% behavioral preservation, and query-time efficiency. All comparisons are paired
% at the example level. We report quality against the dataset reference using
% token-F1 and LLM-judge scores, and output preservation by comparing
% \methodname's output directly against the direct video-in-context
% baseline for the same video and prompt. Judge scores are produced on a 1--5
% scale and rescaled to $[0,1]$.
% \\\\
% For each benchmark and metric, we compute
% \[
%     \Delta =
%     \mathrm{Score}(\methodname)
%     -
%     \mathrm{Score}(\textsc{Direct}).
% \]
% Non-inferiority is declared when the lower bound of the 95\% paired bootstrap
% confidence interval exceeds the pre-declared margin. We use margins of $-0.05$
% for token-F1 and $-0.15$ for rescaled judge score. Equivalence is declared when
% the full confidence interval lies within the corresponding symmetric margin.

\begin{table}[t]
\centering
\resizebox{0.9\columnwidth}{!}{%
\renewcommand{\arraystretch}{1.25}
\begin{tabular}{l  cc  cc}
\toprule
\multirow{2}{*}{\textbf{Subset}}
  & \multicolumn{2}{c}{\textbf{SmolVLM 500M}}
  & \multicolumn{2}{c}{\textbf{SmolVLM 2.2B}} \\
\cmidrule(lr){2-3}\cmidrule(lr){4-5}
& Base & V2L\,($\Delta$) & Base & V2L\,($\Delta$) \\
\midrule
Short caption & 0.629 & 0.535\,(-0.094) & 0.556 & 0.579\,(+0.022) \\
Detailed caption & 0.476 & 0.401\,(-0.074) & 0.526 & 0.463\,(-0.063) \\
Camera & 0.310 & 0.131\,(-0.178) & 0.478 & 0.392\,(-0.085) \\
Background & 0.642 & 0.523\,(-0.117) & 0.588 & 0.606\,(+0.018) \\
Main object & 0.517 & 0.442\,(-0.075) & 0.546 & 0.514\,(-0.032) \\
\bottomrule
\end{tabular}
}
\caption{
  VDC results broken down by caption style.
}
\label{tab:vdc_styles}
\end{table}
\begin{table}[t]
\centering
\resizebox{0.9\columnwidth}{!}{%
\renewcommand{\arraystretch}{1.25}
\begin{tabular}{l  cc  cc}
\toprule
\multirow{2}{*}{\textbf{Subset}}
  & \multicolumn{2}{c}{\textbf{SmolVLM 500M}}
  & \multicolumn{2}{c}{\textbf{SmolVLM 2.2B}} \\
\cmidrule(lr){2-3}\cmidrule(lr){4-5}
& Base & V2L\,($\Delta$) & Base & V2L\,($\Delta$) \\
\midrule
Caption & 0.418 & 0.324\,(-0.094) & 0.465 & 0.400\,(-0.065) \\
Events & 0.201 & 0.169\,(-0.032) & 0.340 & 0.267\,(-0.073) \\
Objects & 0.368 & 0.327\,(-0.043) & 0.457 & 0.392\,(-0.065) \\
Spatial caption & 0.424 & 0.329\,(-0.095) & 0.519 & 0.426\,(-0.094) \\
Temporal caption & 0.260 & 0.242\,(-0.018) & 0.404 & 0.360\,(-0.045) \\
\bottomrule
\end{tabular}
}
\caption{
CaReBench  results  broken down by subset.
}
\label{tab:carebench_subsets}
\end{table}
\begin{table*}[t]
\centering
\resizebox{0.8\textwidth}{!}{%
\renewcommand{\arraystretch}{1.12}
\setlength{\tabcolsep}{3.0pt}
\begin{tabular}{l  r r r r c c  r r r r c c}
\toprule
\multicolumn{13}{l}{\textbf{LLM Judge}} \\
\cmidrule(r){1-13}
\multirow{2}{*}{\textbf{Benchmark}}
  & \multicolumn{6}{c}{\textbf{SmolVLM 500M}}
  & \multicolumn{6}{c}{\textbf{SmolVLM 2.2B}} \\
\cmidrule(lr){2-7}\cmidrule(lr){8-13}
& Base & V2L & $\Delta$ & CI & Eq & NI & Base & V2L & $\Delta$ & CI & Eq & NI \\
\midrule
NExT-QA (open) & 0.501 & 0.547 & +0.046 & [+0.007, +0.084] & \good{Y} & \good{Y} & 0.597 & 0.610 & +0.013 & [-0.022, +0.048] & \good{Y} & \good{Y} \\
ActivityNet-QA & 0.524 & 0.541 & +0.016 & [-0.031, +0.064] & \good{Y} & \good{Y} & 0.627 & 0.531 & -0.096 & [-0.144, -0.049] & \good{Y} & \good{Y} \\
PLM-SGQA & 0.390 & 0.317 & -0.074 & [-0.113, -0.034] & \good{Y} & \good{Y} & 0.493 & 0.295 & -0.198 & [-0.236, -0.161] & \bad{--} & \bad{--} \\
VidCapBench & 0.502 & 0.451 & -0.050 & [-0.071, -0.030] & \good{Y} & \good{Y} & 0.551 & 0.475 & -0.076 & [-0.096, -0.055] & \good{Y} & \good{Y} \\
\midrule
\textit{Average} & 0.487 & 0.460 & -0.027 & [-0.043, -0.011] & \good{Y} & \good{Y} & 0.562 & 0.477 & -0.085 & [-0.101, -0.069] & \good{Y} & \good{Y} \\
\midrule[0.8pt]
\multicolumn{13}{l}{\textbf{Token F1}} \\
\cmidrule(r){1-13}
\multirow{2}{*}{\textbf{Benchmark}}
  & \multicolumn{6}{c}{\textbf{SmolVLM 500M}}
  & \multicolumn{6}{c}{\textbf{SmolVLM 2.2B}} \\
\cmidrule(lr){2-7}\cmidrule(lr){8-13}
& Base & V2L & $\Delta$ & CI & Eq & NI & Base & V2L & $\Delta$ & CI & Eq & NI \\
\midrule
NExT-QA (open) & 0.129 & 0.068 & -0.061 & [-0.076, -0.046] & \bad{--} & \bad{--} & 0.140 & 0.076 & -0.063 & [-0.079, -0.048] & \bad{--} & \bad{--} \\
ActivityNet-QA & 0.197 & 0.023 & -0.174 & [-0.199, -0.149] & \bad{--} & \bad{--} & 0.149 & 0.013 & -0.136 & [-0.156, -0.117] & \bad{--} & \bad{--} \\
PLM-SGQA & 0.081 & 0.225 & +0.145 & [+0.131, +0.158] & \bad{--} & \warn{Y} & 0.092 & 0.203 & +0.111 & [+0.098, +0.124] & \bad{--} & \warn{Y} \\
VidCapBench & 0.216 & 0.209 & -0.007 & [-0.019, +0.004] & \good{Y} & \good{Y} & 0.196 & 0.218 & +0.022 & [+0.010, +0.033] & \good{Y} & \good{Y} \\
\midrule
\textit{Average} & 0.156 & 0.131 & -0.024 & [-0.041, -0.008] & \good{Y} & \good{Y} & 0.144 & 0.128 & -0.017 & [-0.032, -0.002] & \good{Y} & \good{Y} \\
\bottomrule
\end{tabular}
}
\caption{
Comparison of the base model with video and \methodname generated adapters, across video question answering benchmarks using LLM Judge scores and Token F1. We report mean scores, the paired difference $\Delta$ (V2L $-$ Base), 95\% confidence intervals, and the statistical equivalence (Eq) and non-inferiority (NI) criteria.
}
\label{tab:qa_main}
\end{table*}

\subsection{Captioning}

\methodname passes both non-inferiority and equivalence on \textbf{all 10
benchmark--scale combinations} under the LLM judge and all 10 under token-F1
(Table~\ref{tab:cap_main}). For SmolVLM 2.2B, \methodname recovers \textbf{91.9\%} of the base model's judge score, while for SmolVLM 500M, it recovers
\textbf{84.2\%}.

\paragraph{Per-benchmark analysis.}
Recovery rates at 500M span 79--96\%, with compact clip-aligned benchmarks
(PLM-RCap, PLM-RDCap) easiest to internalize and temporally dense benchmarks
(VDC, ActivityNet~Captions) hardest.
Scale narrows this spread considerably: at 2.2B the floor rises to 85\% and
the ceiling breaks above the base, with PLM-RCap \emph{surpassing} the base
outright (CI entirely above zero) and PLM-RDCap reaching de-facto equivalence
(CI straddling zero).
The benchmarks most sensitive to scale---particularly VDC, where the gap
contracts fourfold---are those requiring compression of visually diverse,
longer-form descriptions; benchmarks with consistently structured references
recover well at both scales.

\paragraph{Token F1.}
Token-F1 provides independent reference-based corroboration: the mean paired
delta is $-0.001$ at 500M and $0.000$ at 2.2B.
\methodname exceeds base on 3 of 5 benchmarks at 500M (ActivityNet Captions
${+}0.007$, PLM-RDCap ${+}0.009$, PLM-RCap ${+}0.026$) and 2 of 5 at 2.2B.
The PLM-RCap result at 500M is notable: $+0.026$ ($+14.7\%$; CI $[{+}0.021,
{+}0.031]$) with no token-level supervision.

% \subsection{Output Preservation}

% Table~\ref{tab:preservation} evaluates whether \methodname reproduces
% the direct baseline's own behavior, rather than only matching dataset
% references. This matters because a model could obtain reasonable reference
% scores by producing generic captions. Preservation directly asks whether the
% adapter leads the frozen model to produce outputs similar to those it would have
% produced when seeing the video in context.
% \\\\
% The preservation scores show that \methodname captures a meaningful
% fraction of the base model's video-conditioned behavior. Preservation improves
% with scale on all captioning benchmarks. For example, VDC aggregate similarity
% increases from $0.596$ at 500M to $0.705$ at 2.2B, and the fraction of examples
% with similarity at least $0.8$ increases from $32.6\%$ to $60.9\%$. This
% supports the interpretation that the adapter is not merely learning a captioning
% prior; it preserves video-specific behavior from the direct baseline.
% \\\\
% QA preservation is lower and less scale-sensitive, with mean similarities mostly
% between $0.52$ and $0.65$. This is consistent with the main QA results: the
% adapter preserves enough information to answer some downstream questions, but
% fine-grained query-specific behavior is harder to reproduce than caption-level
% semantics.

% \input{tables/tab3_mean_similarity}

\subsection{Fine-Grained Captioning}

% Tables~\ref{tab:vdc_styles} and~\ref{tab:carebench_subsets} show where the
% adapter loses information. On VDC, \methodname is strongest on broad
% semantic descriptions and background descriptions, but weakest on camera-related
% descriptions. At 2.2B, it slightly improves over the direct baseline on short
% captions and background descriptions, while remaining below the baseline on
% detailed caption, camera, and main-object prompts. The largest degradation is on
% camera descriptions, indicating that viewpoint, framing, and camera motion are
% not preserved as well as event and scene semantics.
% \\\\
% CaReBench shows a similar pattern. The smallest gaps occur on event or temporal
% descriptions, while larger gaps appear on caption and spatial-description
% subsets. This suggests that \methodname better preserves what is
% happening in the video than where every object is located or how the scene is
% spatially arranged. In other words, the adapter captures high-level visual
% semantics more reliably than fine-grained spatial detail.

Tables~\ref{tab:vdc_styles} and~\ref{tab:carebench_subsets} break VDC and
CaReBench into caption styles and semantic dimensions.

\textbf{VDC}
Four of five VDC styles maintain 81--85\% recovery at 500M: short (85.1\%,
$\Delta = {-}0.094$), detailed (84.2\%, $\Delta = {-}0.074$), background
(81.5\%, $\Delta = {-}0.117$), main object (85.5\%, $\Delta = {-}0.075$).
\emph{Camera} captions are the outlier: 
% \methodname-SmolVLM\_500M achieves only
% \textbf{42.3\% recovery} ($\Delta = {-}0.178$; base 0.310, V2L 0.131),
% as cinematographic attributes---shot framing, viewpoint, camera motion---are
% difficult to encode as weight perturbations at 500M scale.
At 500M, \methodname achieves only 42.3\% recovery
($\Delta=-0.178$; base 0.310, V2L 0.131), as cinematographic
attributes---shot framing, viewpoint, and camera motion---are difficult to
encode as weight perturbations at this scale. At 2.2B, \methodname
recovers 82.0\% ($\Delta=-0.085$), a gain of +39.7 pp.
\methodname recovers \textbf{82.0\%} ($\Delta = {-}0.085$), a gain of
\textbf{+39.7 pp}---the largest single-dimension scale improvement in the
fine-grained evaluation---This suggests that part of the camera-description gap is capacity-related,
although targeted camera-motion supervision or adaptive rank may still be needed.
% confirming the camera bottleneck is a capacity
% limitation rather than a fundamental barrier.
At 2.2B, two styles exceed the base outright: short captions (104.1\%,
$\Delta = {+}0.022$) and background (103.1\%, $\Delta = {+}0.018$).

\paragraph{CaReBench}
Temporal captioning is best-recovered at both scales (500M: 93.1\%,
$\Delta = {-}0.018$; 2.2B: 89.1\%, $\Delta = {-}0.045$); objects follow
(500M: 88.9\%; 2.2B: 85.8\%).
Holistic captioning and spatial description are hardest at 500M (77.5\% and
77.6\%), but scale closes the gap strongly: holistic reaches 86.0\%
($+8.5$~pp) and spatial 82.1\% ($+4.5$~pp) at 2.2B.
The events dimension inverts: recovery falls from 84.1\% (500M) to 78.5\%
(2.2B, $-5.6$~pp) as the 2.2B base improves substantially on event enumeration
(base: 0.201 $\to$ 0.340), raising the compression target beyond the adapter's fixed rank.
\begin{figure*}[t]
  \centering
  \resizebox{0.8\textwidth}{!}{%
  \begin{minipage}{\textwidth}
    \centering

    \begin{subfigure}[b]{0.35\textwidth}
      \centering
      \includegraphics[width=\linewidth]{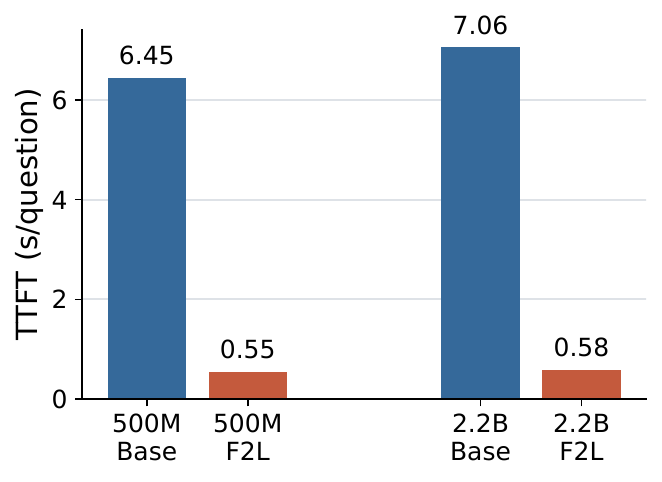}
      \caption{Single-question average TTFT, with time taken to internalize the video accounted.}
      % Frames2LoRA achieves
      %   $11.7\times$ (500M) and $12.2\times$ (2.2B) lower TTFT than
      %   direct inference.
      \label{fig:eff_ttft}
    \end{subfigure}
    \hfill
    \begin{subfigure}[b]{0.55\textwidth}
      \centering
      \includegraphics[width=\linewidth]{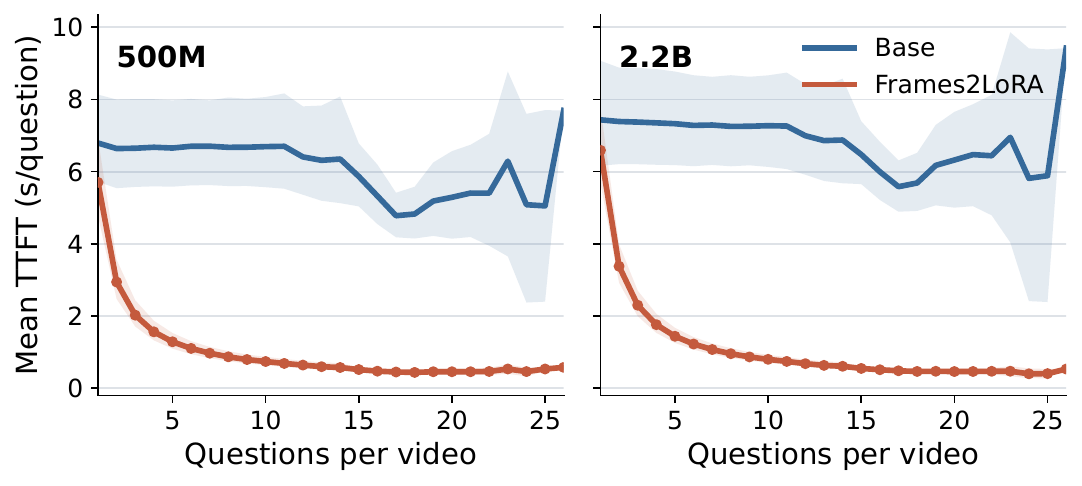}
      \caption{Amortized TTFT per question vs.\ number of questions per video
        (shaded band = bootstrap 95\% confidence interval).}
        
        % The speedup grows from
        % $5.10\times$ at one question to $34.04\times$ at 25 questions per video.
      \label{fig:amortization}
    \end{subfigure}

  \end{minipage}%
  }

  \caption{
    Inference efficiency on VidCapBench, comparing the base model and \methodname.
  }
  \label{fig:efficiency}
\end{figure*}

\subsection{Video Question Answering}

\methodname is trained exclusively on captioning; video QA is entirely a
zero-shot transfer task.
The LLM judge passes non-inferiority and equivalence on \textbf{7 of 8}
benchmark--scale combinations (Table~\ref{tab:qa_main}).
\begin{figure}[t]
    \centering
    \begin{subfigure}{\columnwidth}
        \centering
        \includegraphics[width=\columnwidth]{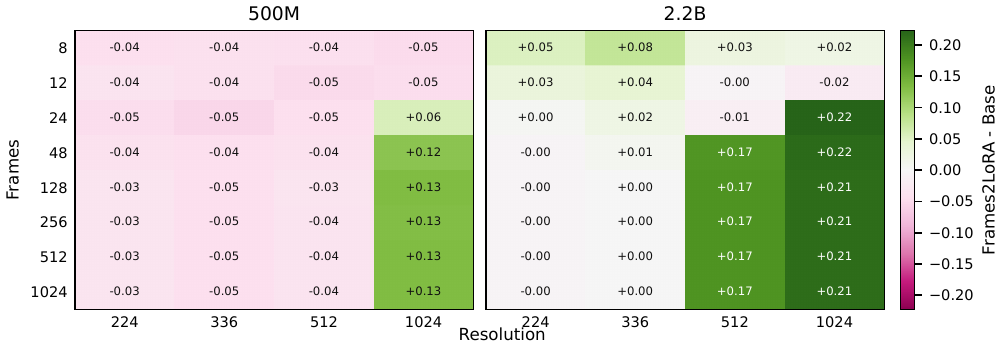}
        \caption{Change in mean Token-F1 from replacing in-context video tokens with \methodname.}
        \label{fig:vdc-scaling-quality}
    \end{subfigure}

    \vspace{0.5em}

    \begin{subfigure}{\columnwidth}
        \centering
        \includegraphics[width=\columnwidth]{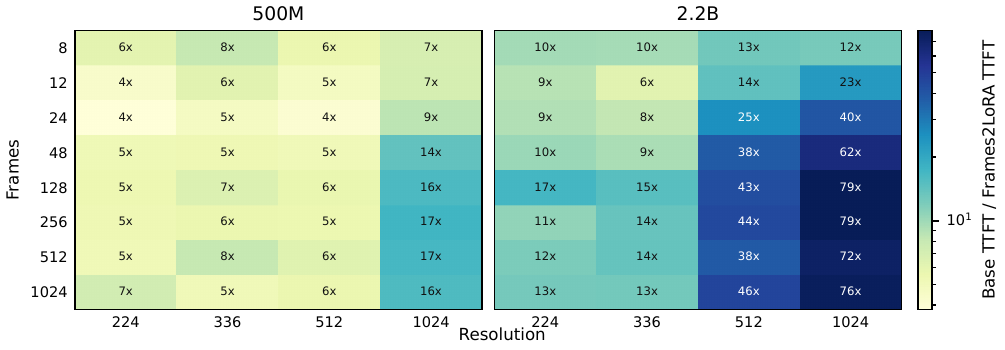}
        \caption{Query-time TTFT speedup of \methodname over the base video-in-context model.}
        \label{fig:vdc-scaling-ttft}
    \end{subfigure}

    \vspace{0.5em}

    \begin{subfigure}{\columnwidth}
        \centering
        \includegraphics[width=\columnwidth]{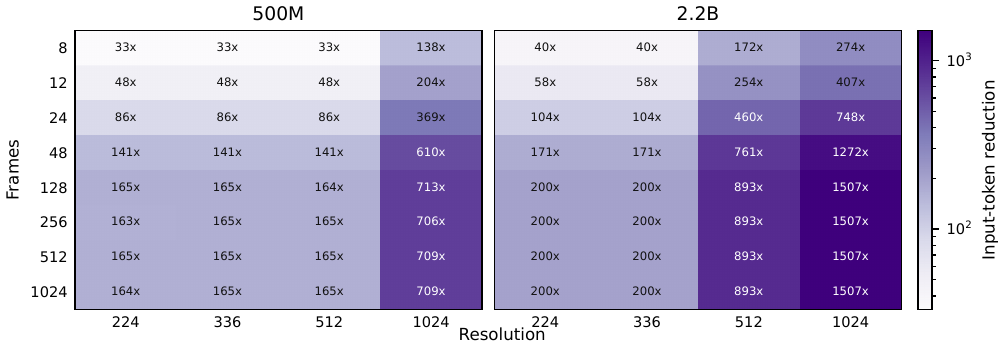}
        \caption{Input-token reduction achieved by \methodname during answering.}
        \label{fig:vdc-scaling-token-reduction}
    \end{subfigure}

    \caption{
    Scaling behavior on VDC background captioning across frame count and spatial resolution.
    }
    \label{fig:vdc-scaling}
\end{figure}

\paragraph{Per-benchmark judge analysis.}
Across the four QA benchmarks, \methodname matches or exceeds the base on two
of four at 500M and one of four at 2.2B, with NExT-QA being the standout:
\methodname \emph{surpasses} the base at both scales, with the 500M CI lying
entirely above zero.
The single failure---PLM-SGQA at 2.2B---is  instructive rather than
representative. The same benchmark passes comfortably at 500M points does not point to a fundamental limitation of parametric QA
internalization.

\paragraph{Token-F1 and the verbosity effect.}
Token-F1 diverges from the judge on short-answer QA. Token-F1 exposes a strong format mismatch on short-answer QA. This does not
necessarily imply semantic failure, but it shows that captioning-trained
\methodname tends to produce more verbose answers than the direct baseline.
% This is a style artefact, not a quality failure.
On ActivityNet-QA, \methodname token-F1 falls to 12\% of base at 500M (0.023
vs.\ 0.197) and 9\% at 2.2B (0.013 vs.\ 0.149); on NExT-QA it is 53\%---yet
both pass the judge test.
The base VLM gives short, often one-to-three-word answers; \methodname, trained
on captioning, generates verbose summaries.
Token-F1 is penalised by both the length mismatch and paraphrase variation,
while the judge evaluates semantic correctness independently of response length.
Two contrasts support this interpretation: PLM-SGQA---with longer, descriptive
references---reverses direction entirely (500M: $\Delta = {+}0.145$; 2.2B:
$\Delta = {+}0.111$); VidCapBench reaches near-parity
($\Delta = {-}0.007$ / ${+}0.022$).

% Table~\ref{tab:qa_main} shows that QA is harder than captioning. Under
% LLM-judge evaluation, \methodname remains close to direct
% video-in-context inference on most benchmark-scale pairs. It matches or exceeds
% the direct baseline on NExT-QA at both model scales and on ActivityNet-QA at
% 500M. It also remains within the non-inferiority and equivalence margins on
% VidCapBench. The main failure is PLM-SGQA at 2.2B, where the judge-score delta
% is outside the margin.
% \\\\
% Token-F1 gives a more mixed picture. It drops sharply on NExT-QA and
% ActivityNet-QA, even when the judge score remains competitive, but improves on
% PLM-SGQA. This divergence suggests that token-F1 is sensitive to answer wording
% and benchmark format, while the judge score better captures semantic correctness
% for open-ended QA. Therefore, the QA results should be interpreted more
% cautiously than the captioning results: \methodname transfers to QA, but
% query-specific retrieval from adapter weights is less reliable than generating a
% general video description.

\subsection{Frame and Resolution Generalization}
\methodname checkpoints were trained with uniform sampling at 12 frames and
384px resolution. We test out-of-distribution scaling on VDC background
captioning by sweeping $\{8,12,24,48,128,256,512,1024\}$ frames and
$\{224,336,512,1024\}$ resolution for both 500M and 2.2B models. We compare
video-in-context inference with \methodname using Token-F1, query-time TTFT (Time to First Token), and
input-token reduction during answering (Fig.~\ref{fig:vdc-scaling}).

Despite being trained at a single frame count-resolution setting, \methodname remains
stable across the sweep. For 500M, \methodname is close to the base model overall,
with an average Token-F1 change of $-0.012$. At 1024px and high frame counts,
however, \methodname outperforms the base model by $+0.12$ to $+0.13$ Token-F1.
This large gain is partly because direct video-in-context inference becomes
unstable in this regime: the base model often degenerates into repetitive or
gibberish generations when significantly large number of visual tokens are supplied.
The efficiency gains grow with video scale. \methodname reduces query TTFT by a
geometric mean of $6.7\times$ for 500M and $20.1\times$ for 2.2B, with maximum
speedups of $17.2\times$ and $79.1\times$, respectively
(Fig.~\ref{fig:vdc-scaling-ttft}). This is explained by the token compression in
Fig.~\ref{fig:vdc-scaling-token-reduction}: \methodname reduces answer-time input
tokens by $150\times$ for 500M and $302\times$ for 2.2B on average, reaching
$713\times$ and $1507\times$ at the largest settings, since it passes \textbf{zero tokens} during inference.

\subsection{Inference Efficiency}

\begin{figure*}[t]
  \centering
  \includegraphics[width=1.5\columnwidth]{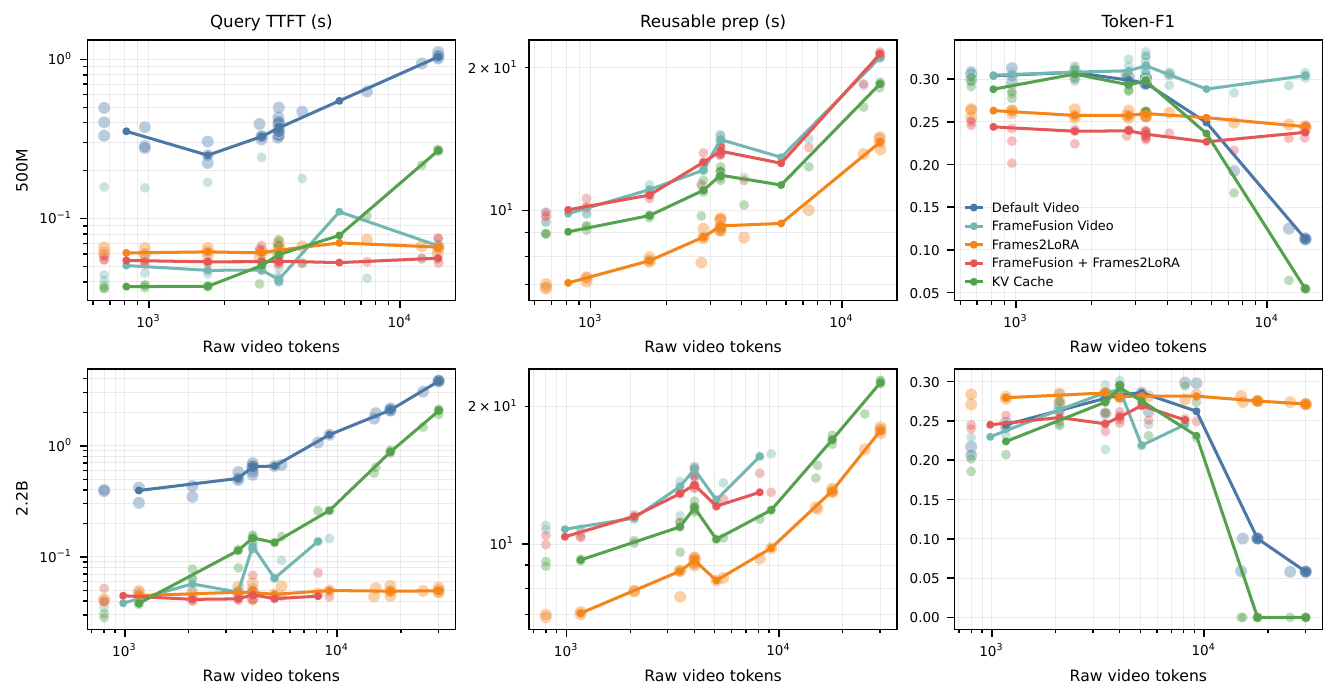}
 \caption{Efficiency comparison across video-token budgets. Columns report query TTFT, reusable preprocessing cost (internalization for \methodname, cache creation for KV Cache, and token compression for FrameFusion), and Token-F1.}
  \label{fig:vdc-token-axis-results-by-model-inset-legend}
\end{figure*}

VidCapBench is a natural setting for evaluating inference efficiency because each
video is associated with multiple questions: in our evaluation split, 100 videos
produce 1,523 total queries, or 15.23 questions per video on average. This
matches the intended use case of \methodname: the video is processed once
to produce a video-specific LoRA, and the adapter is reused for all subsequent
questions about the same video. Thus, unlike direct in-context video inference,
which repeatedly pays the cost of encoding and conditioning on the video,
\methodname pays a one-time setup cost and amortizes it over repeated
queries. Figure~\ref{fig:efficiency} shows this amortization effect on both the 500M and
2.2B backbones. Averaged over all VidCapBench queries, \methodname
reduces TTFT from 6.45s to 0.55s for the 500M model, an $11.75\times$ speedup,
and from 7.06s to 0.58s for the 2.2B model, a $12.11\times$ speedup
(Figure~\ref{fig:eff_ttft}). The prefix-amortization curve in
Figure~\ref{fig:amortization} shows that after 5 questions, amortized TTFT drops
to 1.29s for 500M and 1.44s for 2.2B; after 10 questions, it falls to 0.74s and
0.80s, respectively.

Figure~\ref{fig:vdc-token-axis-results-by-model-inset-legend} further studies
different video inference strategies on 640 samples with varying token counts (by doing using a resolution, frame count grid). We compare \methodname and the default setting with,
FrameFusion \cite{fu2025framefusion} (a token compression technique, compression factor $4$), and KV caching. We also use FrameFusion with \methodname, to show \methodname is compatible with existing token compression techniques. Across token
budgets, \methodname is the only method that provides all three properties needed
for repeated video querying: (1) query TTFT stays nearly constant and low as video
tokens grow, (2) reusable preparation is competitive or fastest and much cheaper
than KV caching at scale, and (3) output quality remains stable as token count increases.
In contrast, the default baseline, token compression results and KV caching scale with token counts. Together, these results show that
\methodname converts video conditioning from a repeated per-query
overhead into a reusable video-specific computation.

\subsection{Chunk Composition}

\begin{figure}[t]
    \centering
    \includegraphics[width=\linewidth]{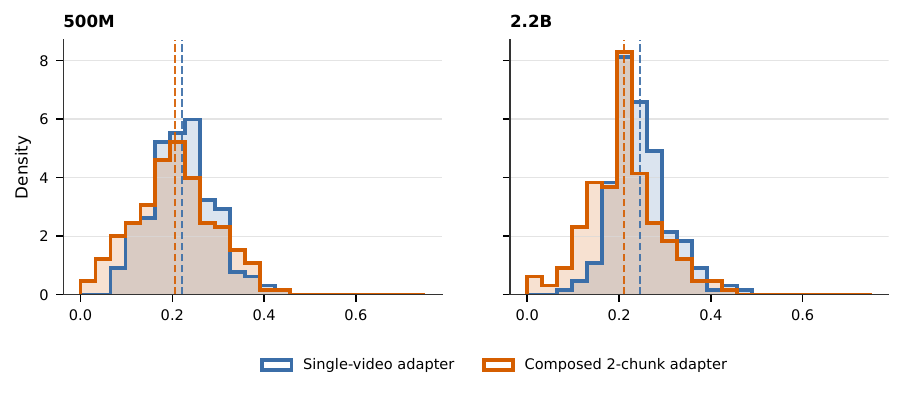}
    \caption{
Two-chunk adapter composition on VDC.
}
    \label{fig:chunking}
\end{figure}

\methodname internalizes a video by generating a LoRA adapter from its visual context. Although the model is trained to produce adapters for single video contexts, the adapter representation admits a simple test-time composition operation: independently internalize two temporal chunks of the same video, concatenate the resulting LoRA ranks, and decode from the composed adapter. We evaluate whether this operation produces coherent video-level generations, rather than degenerate text or captions tied to only one chunk.

We use the VDC short-caption and detailed-caption subsets, with 100 videos from each subset. Each video is split into two equal temporal halves. We compare two conditions: \textit{single-video adapter}, where the full video is internalized as one adapter, and \textit{composed two-chunk adapter}, where the two halves are internalized independently and the resulting adapters are composed before generation. Both conditions use 12 frames per adapter and the same text prompt. Figure~\ref{fig:chunking} shows the resulting token-F1 score distributions against the VDC reference captions. The composed adapter remains close to the single-video adapter at both model scales. 
% For Frames2LoRA-500M, the composed adapter retains 93.1\% of the single-video adapter's mean token-F1, with a mean score of 0.206 compared to 0.221 for the single-video adapter. For Frames2LoRA-2.2B, it retains 86.2\%, with a mean score of 0.211 compared to 0.245.
For \methodname at 500M, the composed adapter retains 93.1\% of the single-video
adapter's mean token-F1, with a mean score of 0.206 compared to 0.221. At 2.2B, it retains 86.2\%, with a mean score of 0.211 compared to 0.245.

% Figure~\ref{fig:chunking} tests whether adapters generated for separate video
% segments can be combined. Each video is split into two temporal halves, each
% half is internalized independently, and the two adapters are concatenated along
% the LoRA rank dimension before generation.
% \\\\
% The composed adapter produces a token-F1 distribution close to the standard
% single-video adapter. This suggests that independently generated LoRA updates
% can be combined without explicit composition training. However, this should be
% interpreted as content aggregation, not full long-video reasoning. Rank-space
% concatenation does not explicitly encode the order of chunks, so temporal
% ordering remains a limitation.

% ---------------------------------------------------------------
\section{Conclusion}

We introduced \textbf{\methodname}, showing that parametric video internalization is achievable: a Perceiver hypernetwork converts a video into a LoRA adapter in a single forward pass, enabling a frozen VLM to answer queries with no visual tokens in context. Trained only on captioning, \methodname is statistically non-inferior and equivalent to direct video-in-context inference across all five captioning benchmarks at both 500M and 2.2B scales, and transfers zero-shot to video QA on seven of eight benchmark-scale pairings.
It remains stable at 1{,}024 frames where direct inference degenerates, achieves 6--76$\times$ lower query latency with up to 1{,}500$\times$ fewer answer-time tokens, and supports rank-space adapter composition for long-video internalization without dedicated training. Across token budgets, \methodname uniquely combines near-constant query TTFT, scalable preprocessing costs below KV caching and token compression, and stable output quality at longer contexts.

\newpage
\section{Limitations}
% ---------------------------------------------------------------

\methodname demonstrates that video context can be internalized into
generated adapter weights, enabling text-only querying after a one-time video
processing step. Our current implementation trains a separate hypernetwork for
each target VLM scale, and we evaluate it on the 500M and 2.2B SmolVLM2
backbones. Extending the same framework to additional VLM families, larger
models, and shared or scale-transferable hypernetworks is an important direction
for future work.

The present training setup uses captioning and summarization supervision. This
makes transfer to video question answering a zero-shot setting, where answer
style can differ from the direct video-in-context baseline. In particular,
\methodname sometimes produces more descriptive answers for short-answer
QA, which can lower lexical-overlap metrics even when the answer is judged
semantically appropriate. Future work can incorporate mixed captioning--QA
supervision, answer-length control, or lightweight calibration for task-specific
formats.

Because \methodname converts a video into a compact adapter, the
representation may emphasize high-level scene and event information over some
fine-grained details. This is most relevant for tasks requiring precise camera,
spatial, or object-level distinctions. Adaptive-rank adapters, richer
internalization objectives, or more targeted supervision may improve preservation
of these details.

Finally, our chunk-composition experiment is an initial two-chunk test. The
result suggests that independently generated adapters can be combined in rank
space, but the current operation does not explicitly model temporal order. More
structured composition mechanisms and audio-visual internalization remain
promising extensions.

\section{Ethics Statement}
Our research does not use any personally identifiable information (PII) and all datasets employed in this work are used in accordance with their respective licenses.

\section*{Acknowledgments}
This research is partially supported by the NVIDIA Academic Grant Program.

% ---------------------------------------------------------------
% \input{sections/ackowledgements}

% TODO: add acknowledgments

\bibliography{custom}

\appendix

\section{LLM Judge Evaluation}
\label{app:judge_prompts}
% ---------------------------------------------------------------

We use an LLM judge for two purposes: reference-based quality scoring and
reference-free output preservation. The judge is
\texttt{Qwen/Qwen3-VL-30B-A3B-Thinking-FP8}, served locally with vLLM through
an OpenAI-compatible API. For the main reported judge scores, we use text-only
judging: the judge receives the task prompt, reference text, and model output,
but no video frames. We set temperature to $0$, use a maximum of 1024 output
tokens for reference-based scoring, and request JSON-formatted outputs. For
pure output similarity, we use the same judge with a maximum of 768 output
tokens.

For reference-based quality, each candidate is scored independently against the
reference. The judge is not shown model names. For auxiliary paired judgments,
the direct baseline and \methodname outputs are anonymized as Candidate
A and Candidate B, and their order is randomized with a fixed seed. These paired
judgments are used as an audit and are not the primary metric unless explicitly
reported.

\subsection{Reference-Based Captioning Judge}

For captioning and description tasks, the judge measures semantic coverage of
the reference caption. Extra details that are absent from the reference are not
penalized unless they directly contradict the reference.

\begin{promptbox}[Captioning Judge Prompt]
\footnotesize
\textbf{Task guidance:} This is video caption/reference evaluation. Judge
semantic coverage, factual precision, and task fit.

\medskip
\texttt{Prompt: \{prompt\}}\\
\texttt{Reference caption: \{reference\}}\\
\texttt{Candidate caption: \{candidate\}}

\medskip
Evaluate only semantic coverage of the reference caption for the requested
caption task. Do not reward or penalize writing style, fluency, verbosity, or
formatting, except when the candidate is invalid or impossible to understand.

Extra details absent from the reference are not automatically wrong. Record
notable extra details separately. Penalize only direct contradictions or missing
reference facts. A direct contradiction requires that the candidate and
reference cannot both be true.

\medskip
\textbf{Coverage scoring rubric:}
5 = complete coverage; 4 = mostly complete; 3 = partial; 2 = weak overlap;
1 = minimal overlap; 0 = unrelated, invalid, nonsensical, or contradicts the
main reference event.

\medskip
Return only valid JSON with fields:
\texttt{coverage\_score} integer 0--5;
\texttt{coverage\_label};
\texttt{covered\_reference\_facts};
\texttt{missing\_reference\_facts};
\texttt{direct\_contradictions};
\texttt{extra\_details};
\texttt{extra\_details\_type};
\texttt{rationale}.
\end{promptbox}

\subsection{Reference-Based QA Judge}

For QA tasks, the judge first extracts the answer implied by the model output
and then compares it to the reference answer. This avoids over-penalizing
verbose outputs that contain the correct answer.

\begin{promptbox}[QA Judge Prompt]
\footnotesize
\textbf{Task guidance:} This is video question answering. Judge semantic answer
correctness. Allow paraphrases and indirect answers. For yes/no references,
infer yes/no from the candidate if the candidate clearly implies it.

\medskip
\texttt{Question: \{question\}}\\
\texttt{Reference answer: \{reference\}}\\
\texttt{Candidate response: \{candidate\}}

\medskip
First extract the candidate's answer to the question, then compare that
extracted answer to the reference. Do not require the candidate to be concise.
A verbose response can be correct if it contains or clearly
gives the answer. If the candidate describes the same scene but does not
answer the requested attribute, action, location, or count, score it as
low-to-partial rather than correct.

\medskip
\textbf{Scoring rubric:}
5 = fully correct; 4 = correct main answer with minor missing specificity or
harmless extra detail; 3 = partially correct; 2 = related but does not clearly
answer; 1 = minimal overlap; 0 = contradiction, different answer, or invalid.

\medskip
Return only valid JSON with fields:
\texttt{extracted\_answer};
\texttt{score} integer 0--5;
\texttt{answer\_label};
\texttt{contains\_answer};
\texttt{direct\_contradiction};
\texttt{extra\_details\_affect\_score};
\texttt{missing\_key\_answer\_parts};
\texttt{contradictory\_parts};
\texttt{rationale}.
\end{promptbox}

\section{Evaluation Prompts and Task Templates}
\label{sec:appendix_prompts}

This appendix provides the exact evaluation prompts and task-specific templates used across all the benchmarks in our experiments. 

% \subsection{Video Internalization Prompt}
% For models utilizing the Frames2LoRA framework, the video is first processed and encoded into the model's context using the following internalization instruction:

% \begin{promptbox}[Internalization Prompt]
%     \texttt{"Internalize this video for later captioning."}
% \end{promptbox}

\subsection{Video Captioning and Description Benchmarks}
Below are the prompts used to generate descriptions for whole videos, clips, and specific features (e.g., spatial layout, temporal progression, and cinematography style).

\begin{promptbox}[ActivityNet Captions]
        \texttt{"Describe what is happening in this video."}
\end{promptbox}

\begin{promptbox}[PLM-RDCap]
    \texttt{"Describe what happens in this video."}
\end{promptbox}

\begin{promptbox}[PLM-RCap]
    \texttt{"Describe what happens in this video clip."}
\end{promptbox}

\begin{promptbox}[Video Description Corpus (VDC)]
\textbf{Short Caption:} \\
\texttt{"Summarize this video in one detailed sentence."}\\\\
\textbf{Detailed Caption:} \\
\texttt{"Describe this video in detail."}\\\\
\textbf{Camera:} \\
\texttt{"Describe the camera work, framing, and viewpoint in this video."}\\\\
\textbf{Background:} \\
\texttt{"Describe the background, setting, and environment in this video."}\\\\
\textbf{Main Object:} \\
\texttt{"Describe the main subject and its actions in this video."}
\end{promptbox}

\begin{promptbox}[CaReBench]
\textbf{Caption:} \\
\texttt{"Describe the video in as much useful visual detail as possible. Include the main activity, visible people or objects, scene context, appearance, and any important visual details that help explain what is happening."}
\\\\
\textbf{Events:} \\
\texttt{"Describe the key visible events in chronological order. Include all important actions and changes you can observe, with enough detail to distinguish each event clearly."}
\\\\
\textbf{Objects:} \\
\texttt{"Describe the important visible objects and entities in the video in as much useful detail as possible. Include their appearance, location, and role in the scene when visible."}
\\\\
\textbf{Spatial Caption:} \\
\texttt{"Describe the spatial layout in as much useful detail as possible: where the people, objects, and scene elements are located, how they are positioned relative to each other, and what parts of the scene are in the foreground, background, left, right, center, above, or below."}
\\\\
\textbf{Temporal Caption:} \texttt{"Describe the temporal progression in as much useful detail as possible. Explain what happens over time, the order of visible actions, and how the scene or subjects change from the beginning to the end."}
\end{promptbox}

\subsection{Video Question Answering (QA) Benchmarks}
For question answering tasks, templates are structured to format the inputs and instructions depending on whether choices are provided (offered options) or hidden.

\begin{promptbox}[NExT-QA]
        \texttt{\{question\}}\\
        Answer only the question, in one sentence.
\end{promptbox}

\begin{promptbox}[ActivityNet-QA]
    \texttt{\{question\}}\\
    Answer only the question, in one sentence.
\end{promptbox}

\begin{promptbox}[PLM-SGQA]
        \texttt{\{question\}}\\
        Answer only the question, in one sentence.
\end{promptbox}

\begin{promptbox}[VidCapBench]
    \texttt{\{question\}\\
    Answer only the question, in one sentence.}
\end{promptbox}

\section{Rank-Direction Ablation}
\label{app:rank_ablation}

\subsection{Setup}

We test whether different rank directions in a generated LoRA adapter contribute
unequally to captioning performance. The ablation is run on 500 examples from
the ActivityNet Captions evaluation split~\citep{krishna2017activitynet}, using
the 2.2B \methodname checkpoint. For each
example, we generate the video-conditioned rank-$16$ adapter and decompose it
into rank-slice pairs $\{(A_r,B_r)\}_{r=1}^{16}$, where
$A_r \in \mathbb{R}^{1 \times d_{\mathrm{in}}}$ and
$B_r \in \mathbb{R}^{1 \times d_{\mathrm{out}}}$. Under our row-vector
implementation, rank slice $r$ contributes
\[
    \Delta y_r = s\,(xA_r^\top)B_r .
\]
We score each slice by the Frobenius norm product
\[
    \|A_r\|_F \cdot \|B_r\|_F .
\]

We evaluate four selection strategies across budgets
$k \in \{1,2,4,8,16\}$:
\begin{itemize}
    \item \textbf{Top-$k$:} retain the $k$ highest-scoring rank slices.
    \item \textbf{Bottom-$k$:} retain the $k$ lowest-scoring rank slices.
    \item \textbf{Random-$k$:} retain $k$ randomly selected slices, averaged over 3 seeds.
    \item \textbf{Remove-Top-$k$:} remove the $k$ highest-scoring slices and retain the remaining $16-k$.
\end{itemize}
We report Token-F1 against reference captions with 95\% bootstrap confidence
intervals over examples.

\begin{table*}[h]
\centering
% \resizebox{\columnwidth}{!}{%
\begin{tabular}{lcccc}
\toprule
$k$ & \textbf{Top-$k$} & \textbf{Bottom-$k$} & \textbf{Random-$k$} & \textbf{Remove-Top-$k$} \\
\midrule
0 (Zero) & 0.0561 \scriptsize{[.052,.060]} & 0.0561 & 0.0561 & 0.0561 \\
1        & 0.0894 \scriptsize{[.083,.096]} & 0.0556 \scriptsize{[.052,.060]} & 0.0709 \scriptsize{[.068,.074]} & \textbf{0.1317} \scriptsize{[.123,.141]} \\
2        & 0.1097 \scriptsize{[.102,.118]} & 0.0662 \scriptsize{[.062,.071]} & 0.0712 \scriptsize{[.069,.074]} & 0.1277 \scriptsize{[.118,.137]} \\
4        & 0.1196 \scriptsize{[.111,.128]} & 0.0803 \scriptsize{[.074,.086]} & 0.0991 \scriptsize{[.095,.103]} & 0.1275 \scriptsize{[.119,.137]} \\
8        & 0.1264 \scriptsize{[.118,.135]} & 0.1128 \scriptsize{[.104,.121]} & 0.1215 \scriptsize{[.117,.126]} & 0.1128 \scriptsize{[.104,.121]} \\
16 (Full)& 0.1262 \scriptsize{[.117,.136]} & 0.1262 & 0.1262 & 0.0561 \\
\bottomrule
\end{tabular}
% }
\caption{Token F1 scores under rank-direction ablation on ActivityNet Captions. Brackets denote 95\% confidence intervals. \emph{Full Adapter} ($k=16$) and \emph{Zero Adapter} ($k=0$) serve as upper and lower baselines.}
\label{tab:rank_ablation_f1}
\end{table*}
\begin{figure}[t]
    \centering
    \includegraphics[width=\columnwidth]{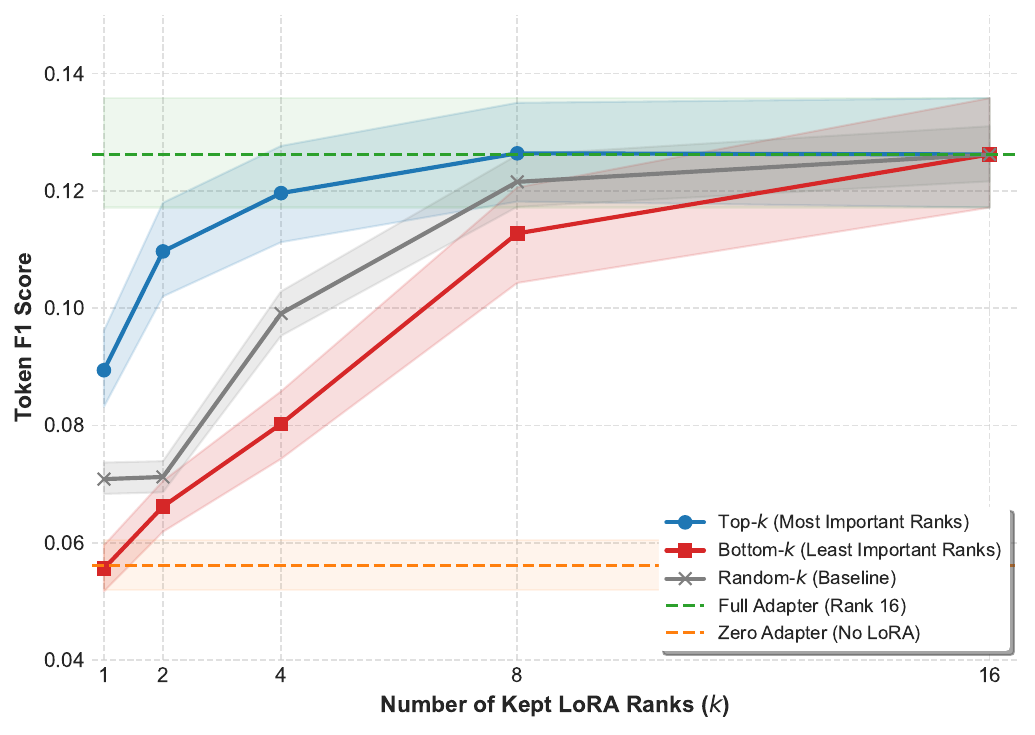}
    \caption{Rank-direction ablation on ActivityNet Captions. Top-$k$ rank slices recover performance faster than random or bottom-$k$ slices, suggesting that the Frobenius norm product is a useful heuristic for rank importance. The Remove-Top-$k$ curve has a higher point estimate than the full adapter at small $k$, but this should be interpreted cautiously because confidence intervals overlap.}
    \label{fig:rank_ablation}
\end{figure}

\subsection{Analysis}

Table~\ref{tab:rank_ablation_f1} reports the numerical ablation results, and
Figure~\ref{fig:rank_ablation} visualizes the same rank-pruning trajectories.

\paragraph{Rank directions are redundant but not exchangeable.}
The generated adapters are compressible along the rank dimension. Retaining the
top-8 rank slices gives a Token-F1 of 0.1264, close to the full rank-16 adapter
score of 0.1262. At $k=4$, the top-$k$ adapter reaches 0.1196, which is 94.8\%
of the full adapter's absolute Token-F1 and recovers 90.6\% of the improvement
over the zero-adapter baseline. This suggests that much of the useful adaptation
is concentrated in a subset of rank directions.

\paragraph{Norm product is a useful heuristic for rank importance.}
The Frobenius norm product separates useful from less useful directions. At
$k=1$, Top-$k$ reaches 0.0894, while Bottom-$k$ reaches 0.0556, slightly below
the zero-adapter baseline of 0.0561. Random-$k$ generally falls between Top-$k$
and Bottom-$k$ at matched budgets. Thus, high-norm rank slices tend to be more
useful, although the norm product should be treated as a heuristic rather than
a complete causal explanation.

\paragraph{Removing the dominant direction has a higher point estimate.}
Removing the highest-norm rank slice gives a higher point estimate than the full
adapter, increasing Token-F1 from 0.1262 to 0.1317. Removing the top four slices
also remains close to the full adapter at 0.1275. Since the confidence intervals
overlap, we treat this as suggestive rather than conclusive. One possible
explanation is that the dominant direction captures a generic captioning prior,
and removing it shifts generation toward more video-specific directions.

\paragraph{Rank ordering is stable across examples.}
The rank ordering is highly consistent across the 500 examples: rank direction
R11 is the highest-scoring direction in all examples, while R7 is consistently
among the lowest-scoring directions. This suggests that the hypernetwork learns
a stable output coordinate system for rank directions, rather than assigning
importance arbitrarily for each video.

\section{Interpreting Hypernetwork-Generated Adapters}
\label{app:interpretability}
\begin{figure*}[t]
    \centering
    \includegraphics[width=\textwidth]{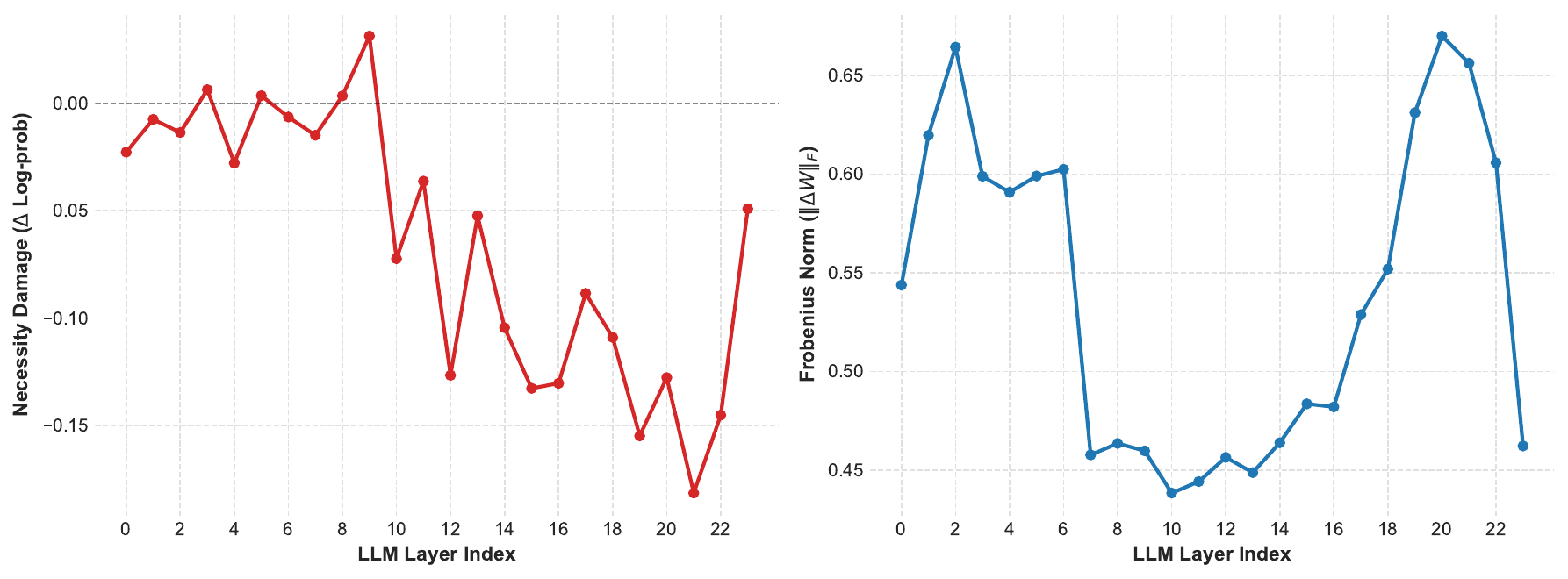}
    \caption{Layer-wise adapter-removal diagnostic. \textbf{Left:} signed removal effect from zeroing one layer's generated LoRA update; negative values indicate that removing the layer lowers the score. \textbf{Right:} Frobenius norm $\|\Delta W\|_F$ of generated LoRA weights across layers.}
    \label{fig:layer_causality}
\end{figure*}
\begin{figure}[t]
    \centering
    \includegraphics[width=\columnwidth]{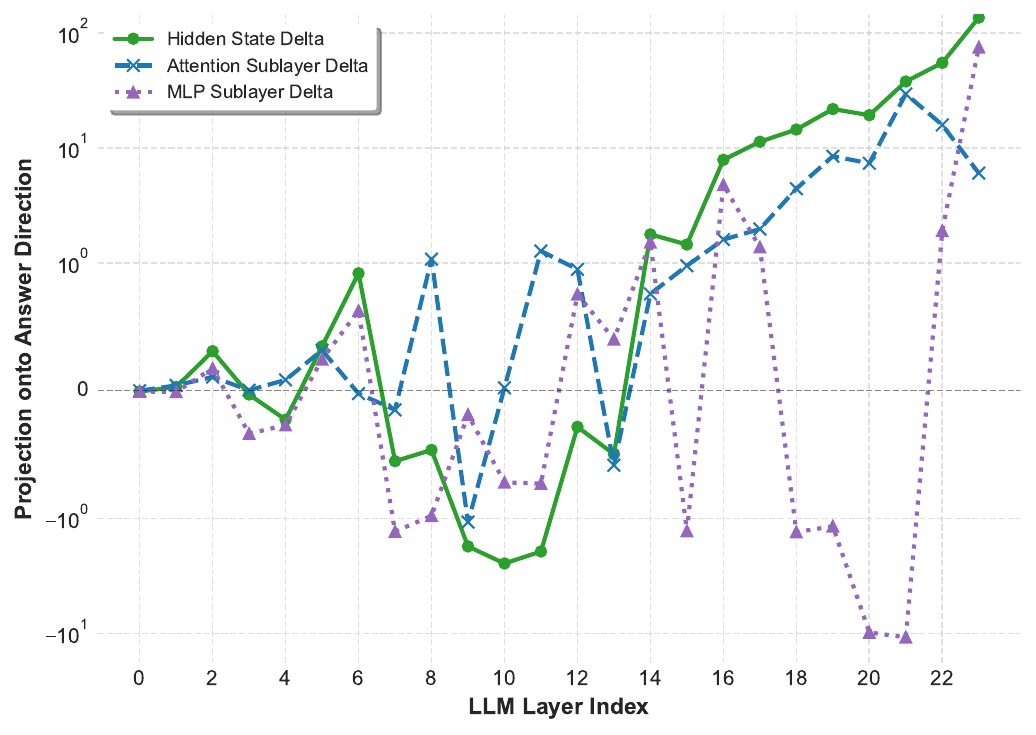}
    \caption{Direct logit attribution of adapter-induced representation shifts projected onto the diagnostic answer direction across 24 LLM layers. Later layers show the largest alignment with the answer direction, suggesting late-layer logit steering.}
    \label{fig:dla}
\end{figure}
\subsection{Setup}
We use two diagnostic interventions to study how generated adapters affect the
frozen 2.2B answer model: layer-wise adapter removal and direct logit
attribution. The experiments are run on CareBench diagnostic examples, including
caption and spatial-caption prompts.
\\\\
Each example is scored by teacher-forced log-probability under the frozen answer
model with the generated adapter active. Since these diagnostics use open-ended
reference strings, we score each reference string and use the highest-scoring
reference for the diagnostic. Candidate strings may contain multiple tokens, so
we score a candidate string $z$ by length-normalized teacher-forced
log-probability:
\begin{equation}
    \ell(z \mid p)
    =
    \frac{1}{|z|}
    \sum_{t=1}^{|z|}
    \log P(z_t \mid z_{<t}, p).
    \label{eq:cand_score}
\end{equation}
The scalar diagnostic score is therefore
\begin{equation}
    \mathcal{S}
    =
    \max_{r \in \mathcal{R}} \ell(r \mid p),
    \label{eq:open_score}
\end{equation}
where $\mathcal{R}$ is the set of reference strings for the example.
\\\\
For direct logit attribution, we need a direction in the output-embedding space.
We use the mean output embedding of the selected reference tokens and denote the
normalized direction by $\hat{d}$. This gives a single diagnostic direction
toward the reference answer/caption.

\subsection{Layer-Wise Adapter Removal}
For each transformer layer $\ell$, we zero out only the generated LoRA update at
that layer and recompute the diagnostic score. We report the signed removal
effect
\begin{equation}
    \mathrm{Effect}_{\ell}
    =
    \mathcal{S}_{\mathrm{without}\ \ell}
    -
    \mathcal{S}_{\mathrm{full}} .
    \label{eq:layer_effect}
\end{equation}
Negative values indicate that removing the layer lowers the score, so the
layer's adapter update is useful under this diagnostic. Values near zero
indicate little measurable effect from removing that layer.
\\\\
Figure~\ref{fig:layer_causality} shows a mismatch between generated-weight norm
and functional effect. Some early layers receive relatively large LoRA updates,
but removing them changes the diagnostic score only weakly. In contrast, several
later layers produce larger negative removal effects, indicating that their
adapter updates matter more for the scored prediction. This suggests that the
adapter is not used uniformly across the transformer stack: early updates may
shape intermediate representations, while later updates appear more directly
connected to the final answer/caption likelihood.
\\\\
This also shows that Frobenius norm alone is not a complete measure of adapter
importance. Large generated weights can be weakly causal under this intervention,
whereas smaller or comparable later-layer updates can have stronger effects on
the output score. We therefore interpret the result as a norm--function
dissociation, not as a full causal explanation of the adapter mechanism.

\subsection{Direct Logit Attribution}
We next ask where the adapter-induced representation shift becomes aligned with
the diagnostic target direction. Let
\[
    \Delta x_\ell
    =
    x^{\mathrm{adapter}}_\ell
    -
    x^{\mathrm{base}}_\ell
\]
denote the residual-stream shift at layer $\ell$, and let $\Delta a_\ell$ and
$\Delta m_\ell$ denote the corresponding attention and MLP sublayer shifts. We
project these shifts onto the diagnostic answer direction:
% \begin{align}
%     \mathrm{DLA}_{\ell}
%     &= \Delta x_\ell \cdot \hat{d}, \\
%     \mathrm{DLA}^{\mathrm{attn}}_{\ell}
%     &= \Delta a_\ell \cdot \hat{d}, \\
%     \mathrm{DLA}^{\mathrm{MLP}}_{\ell}
%     &= \Delta m_\ell \cdot \hat{d}.
% \end{align}
\begin{equation}
\begin{aligned}
    \mathrm{DLA}_{\ell}
    &= \Delta x_\ell \cdot \hat{d}, \\
    \mathrm{DLA}^{\mathrm{attn}}_{\ell}
    &= \Delta a_\ell \cdot \hat{d}, \\
    \mathrm{DLA}^{\mathrm{MLP}}_{\ell}
    &= \Delta m_\ell \cdot \hat{d}.
\end{aligned}
\end{equation}
Figure~\ref{fig:dla} shows that the adapter-induced shift is weakly aligned with
the diagnostic direction in early and middle layers, but becomes much more
aligned in later layers. This matches the layer-removal result: the adapter's
effect becomes most visible close to the output logits.
\\\\
The sublayer breakdown suggests that both attention and MLP components
contribute to this late-stage steering. Rather than claiming that the generated
adapter implements a specific memory mechanism, we interpret the pattern more
conservatively: \methodname appears to induce representation changes
throughout the network, but the changes most directly aligned with the target
answer/caption emerge in later layers.

% 500m trained for 9k steps, 4 A100s, 37 hours

% 2.2b trained for 7k steps, 6 A100s, 201 hours

% hyperparams as follows

% both
% rank 16 
% frames 12
% max video dimension 384 px
% latent size 512
% lr 0.0001
% warmup ratio 0.03
% weight decay 0.01

% 2.2b
% gradient accum steps 5
% per device batch size 8

% 500m
% per device batch size 48
% gradient accumulation steps 2

\section{Training Details}
\label{app:training_details}

Table~\ref{tab:training_details} summarizes the main training configuration for
the two \methodname model scales. In both runs, only the hypernetwork
parameters are trained; the video encoder and answer model remain frozen.

\begin{table}[t]
\centering
\small
\begin{tabular}{lcc}
\toprule
\textbf{Setting} & \textbf{500M} & \textbf{2.2B} \\
\midrule
Training steps & 9{,}000 & 7{,}000 \\
GPUs & 4$\times$A100 & 6$\times$A100 \\
Wall-clock training time & 37 hours & 201 hours \\
Per-device batch size & 48 & 8 \\
Gradient accumulation steps & 2 & 5 \\
Effective batch size & 384 & 240 \\
\midrule
LoRA rank & 16 & 16 \\
Sampled frames & 12 & 12 \\
Max video dimension & 384 px & 384 px \\
Perceiver latent size & 512 & 512 \\
Learning rate & $1\times10^{-4}$ & $1\times10^{-4}$ \\
Warmup ratio & 0.03 & 0.03 \\
Weight decay & 0.01 & 0.01 \\
\bottomrule
\end{tabular}
\caption{Training configuration for the 500M and 2.2B \methodname runs.
Wall-clock training time reports elapsed training time, not total GPU-hours.
Effective batch size is computed as number of GPUs $\times$ per-device batch
size $\times$ gradient accumulation steps.}
\label{tab:training_details}
\end{table}

Both models use rank-$16$ generated LoRA adapters, 12 uniformly sampled frames,
a maximum video dimension of 384 pixels, Perceiver latent size 512, learning
rate $1\times10^{-4}$, warmup ratio 0.03, and weight decay 0.01. The 500M model
is trained for 9{,}000 steps on 4 A100 GPUs for 37 wall-clock hours, with
per-device batch size 48 and gradient accumulation 2, giving an effective batch
size of 384. The 2.2B model is trained for 7{,}000 steps on 6 A100 GPUs for
201 wall-clock hours, with per-device batch size 8 and gradient accumulation 5,
giving an effective batch size of 240.

\section{Additional Results}
\label{sec:results-continued}

\subsection{Distribution Plots}
\label{sec:results-continued-distributions}

\begin{figure*}[t]
  \centering
  \includegraphics[width=\textwidth]{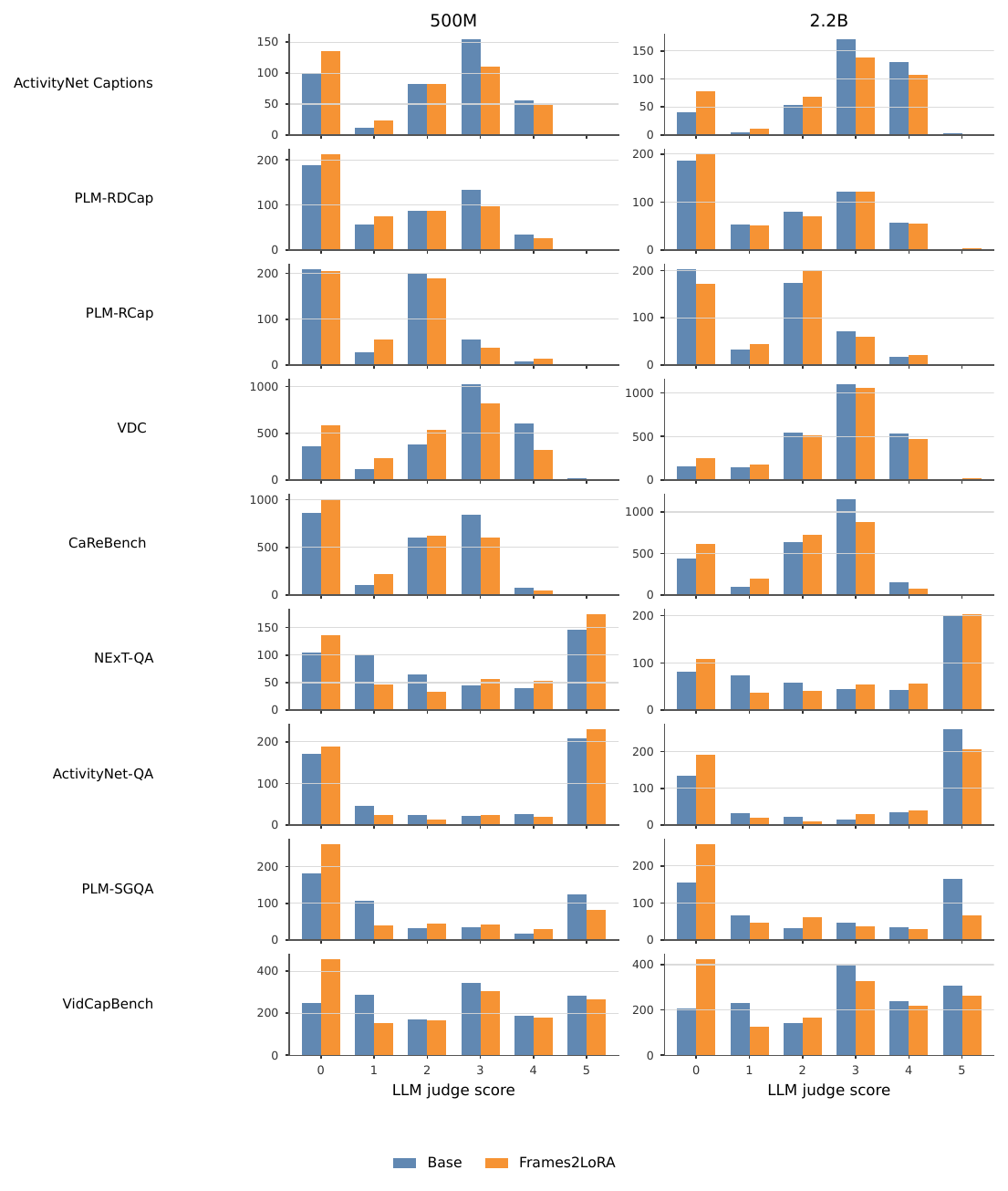}
  \caption{LLM-judge score distributions for the direct baseline and \methodname.}
  \label{fig:appendix-judge-score-distributions-high-level}
\end{figure*}

\begin{figure*}[t]
  \centering
  \includegraphics[width=\textwidth]{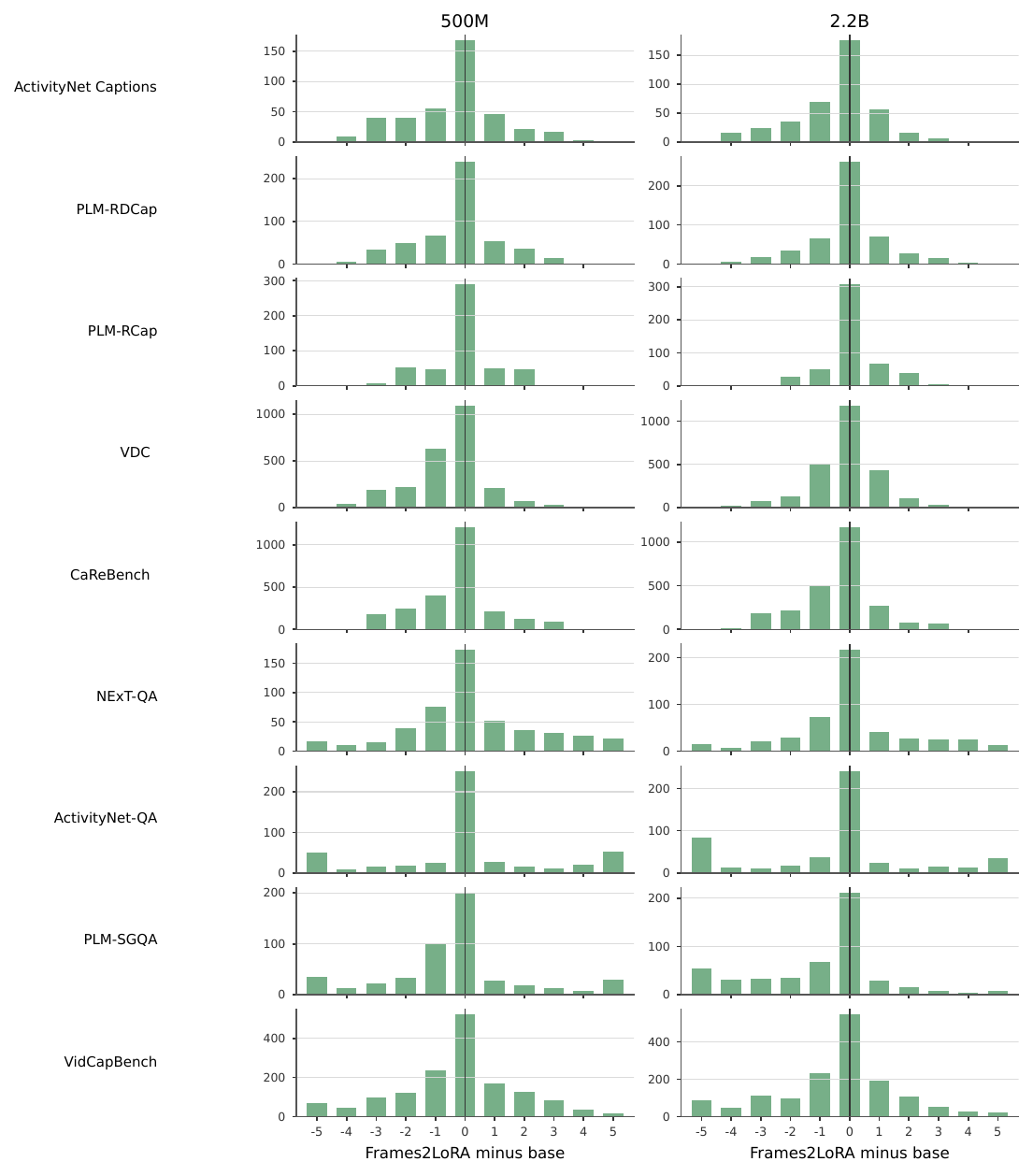}
  \caption{Per-example LLM-judge score differences between \methodname and the direct baseline.}
  \label{fig:appendix-judge-difference-distributions-high-level}
\end{figure*}

\begin{figure*}[t]
  \centering
  \includegraphics[width=\textwidth]{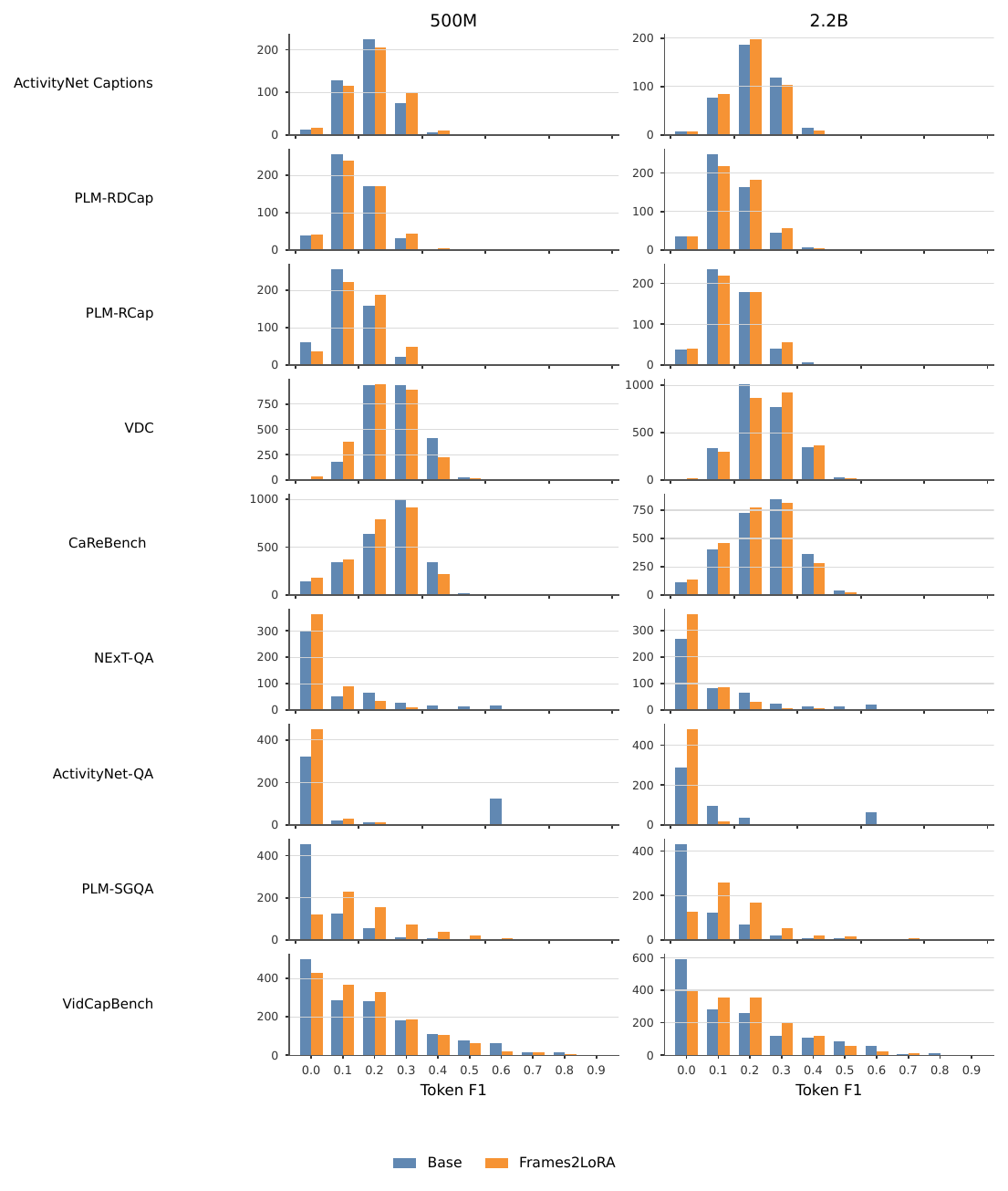}
  \caption{Token-F1 distributions for the direct baseline and \methodname.}
  \label{fig:appendix-token-f1-distributions-high-level}
\end{figure*}

\begin{figure*}[t]
  \centering
  \includegraphics[width=\textwidth]{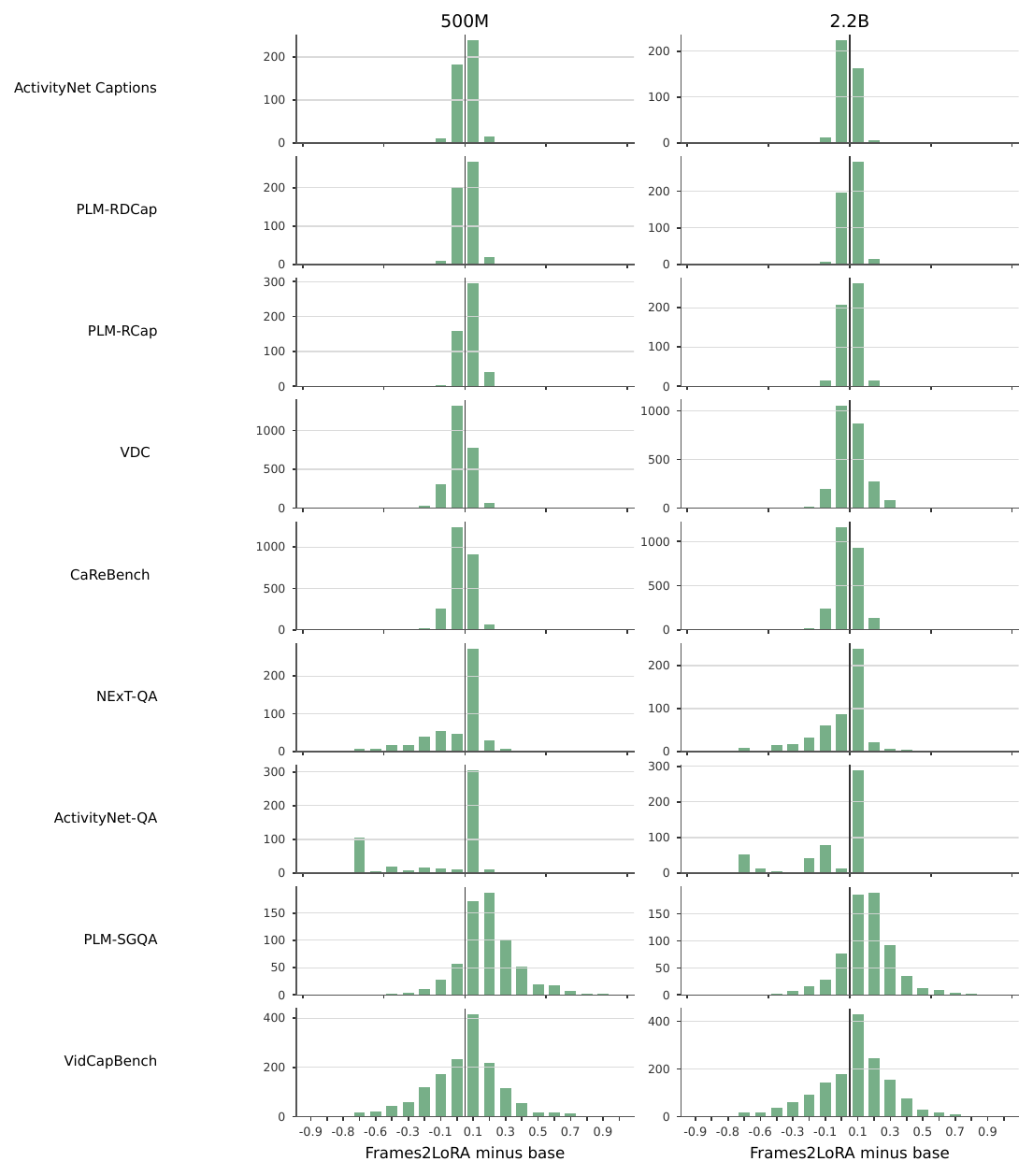}
  \caption{Per-example token-F1 differences between \methodname and the direct baseline.}
  \label{fig:appendix-token-f1-difference-distributions-high-level}
\end{figure*}

Figures~\ref{fig:appendix-judge-score-distributions-high-level}
and~\ref{fig:appendix-judge-difference-distributions-high-level} show the
LLM-judge score distributions and per-example score differences.
Figures~\ref{fig:appendix-token-f1-distributions-high-level}
and~\ref{fig:appendix-token-f1-difference-distributions-high-level} show the
corresponding token-F1 distributions and differences.

\subsection{Spider Plots}
\label{sec:results-continued-spider-plots}
\begin{figure*}[t]
  \centering
  \includegraphics[width=0.88\textwidth]{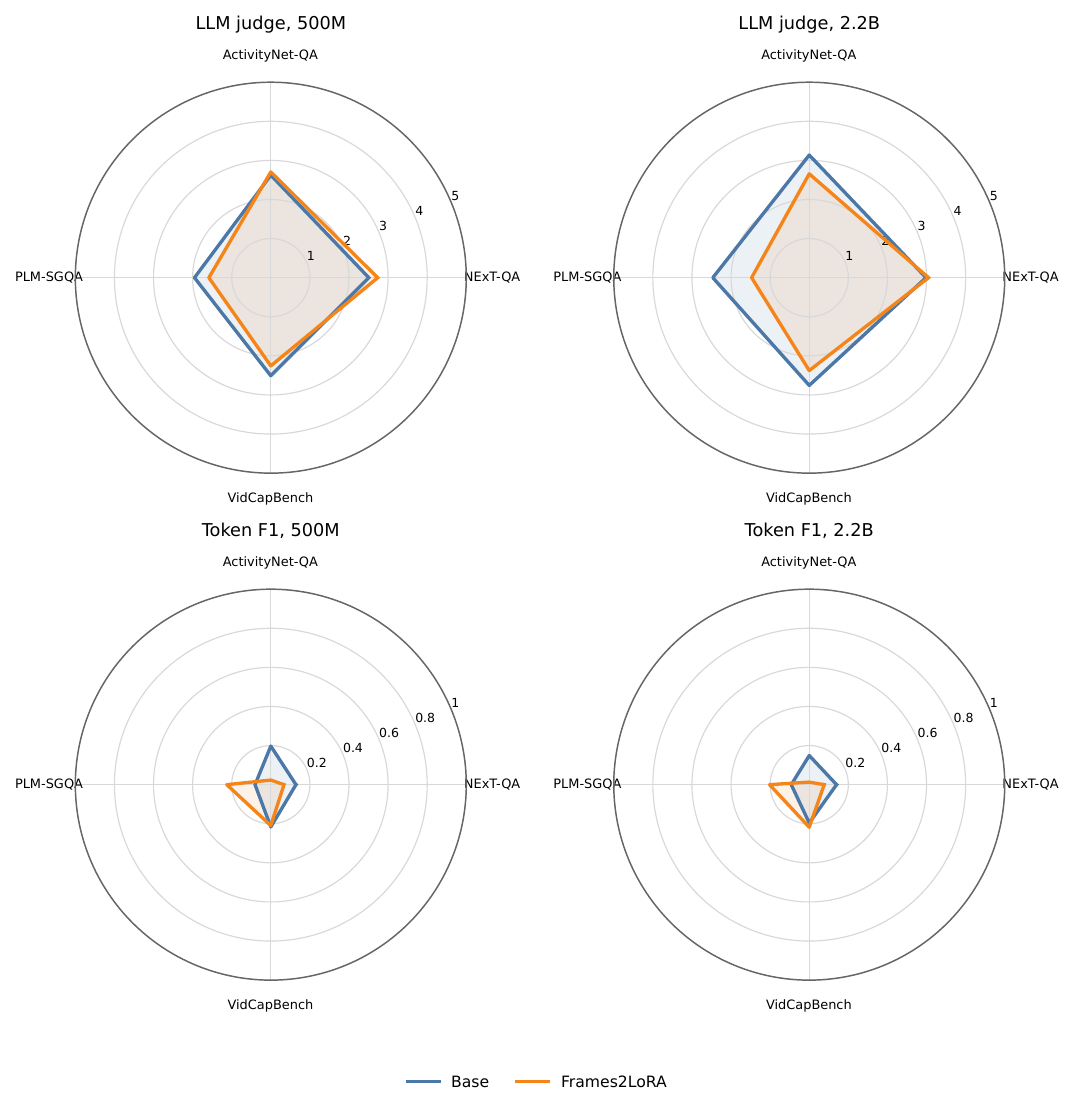}
  \caption{Spider plot for video question answering benchmarks.}
  \label{fig:appendix-spider-qa}
\end{figure*}

\begin{figure*}[t]
  \centering
  \includegraphics[width=0.88\textwidth]{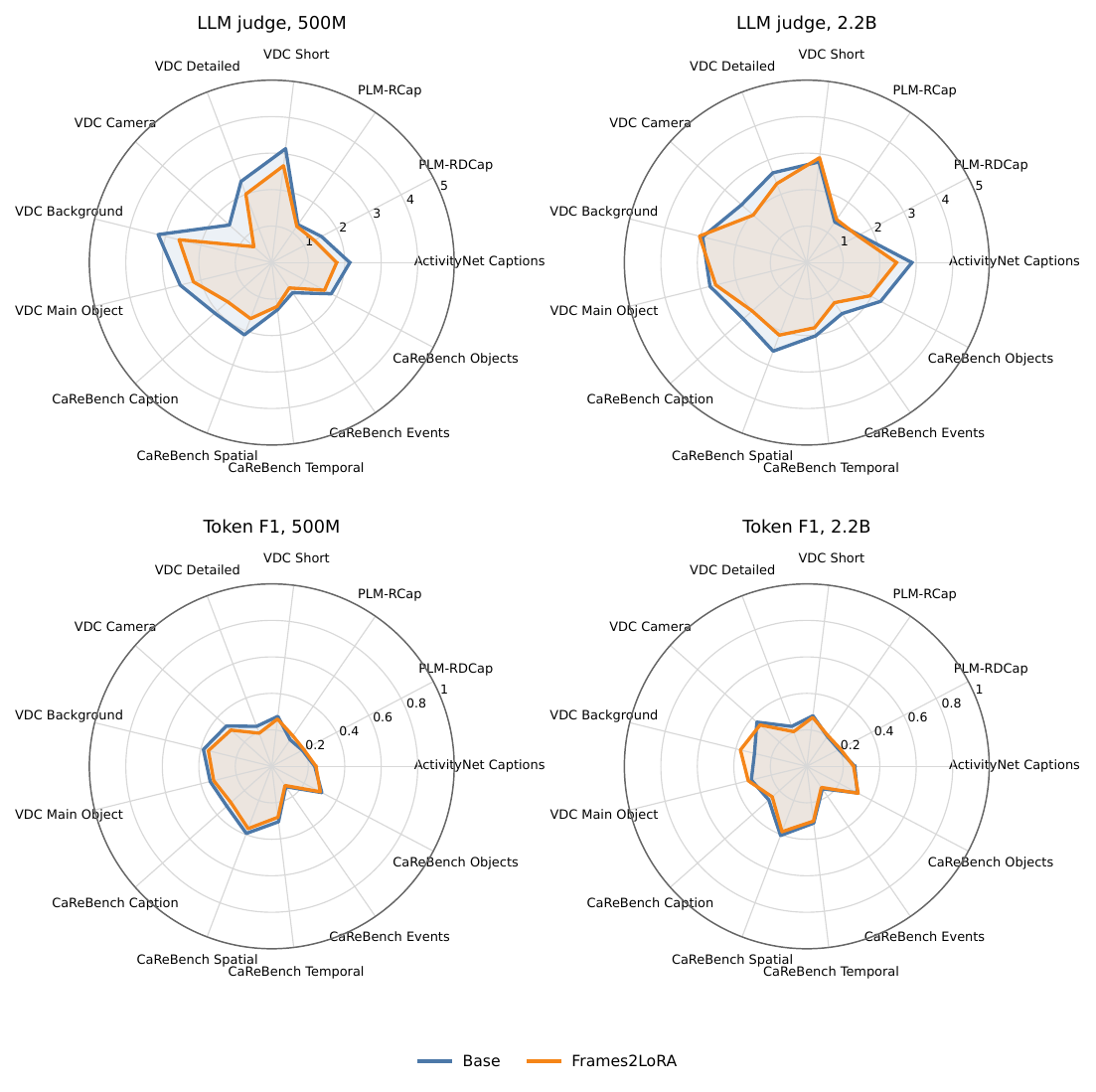}
  \caption{Spider plot for video captioning benchmarks.}
  \label{fig:appendix-spider-captioning-granular}
\end{figure*}

Figures~\ref{fig:appendix-spider-qa}
and~\ref{fig:appendix-spider-captioning-granular} show the QA and captioning
spider plots.

\section{Qualitative Examples}

\begin{figure*}[t]
  \centering
  \includegraphics[width=\linewidth]{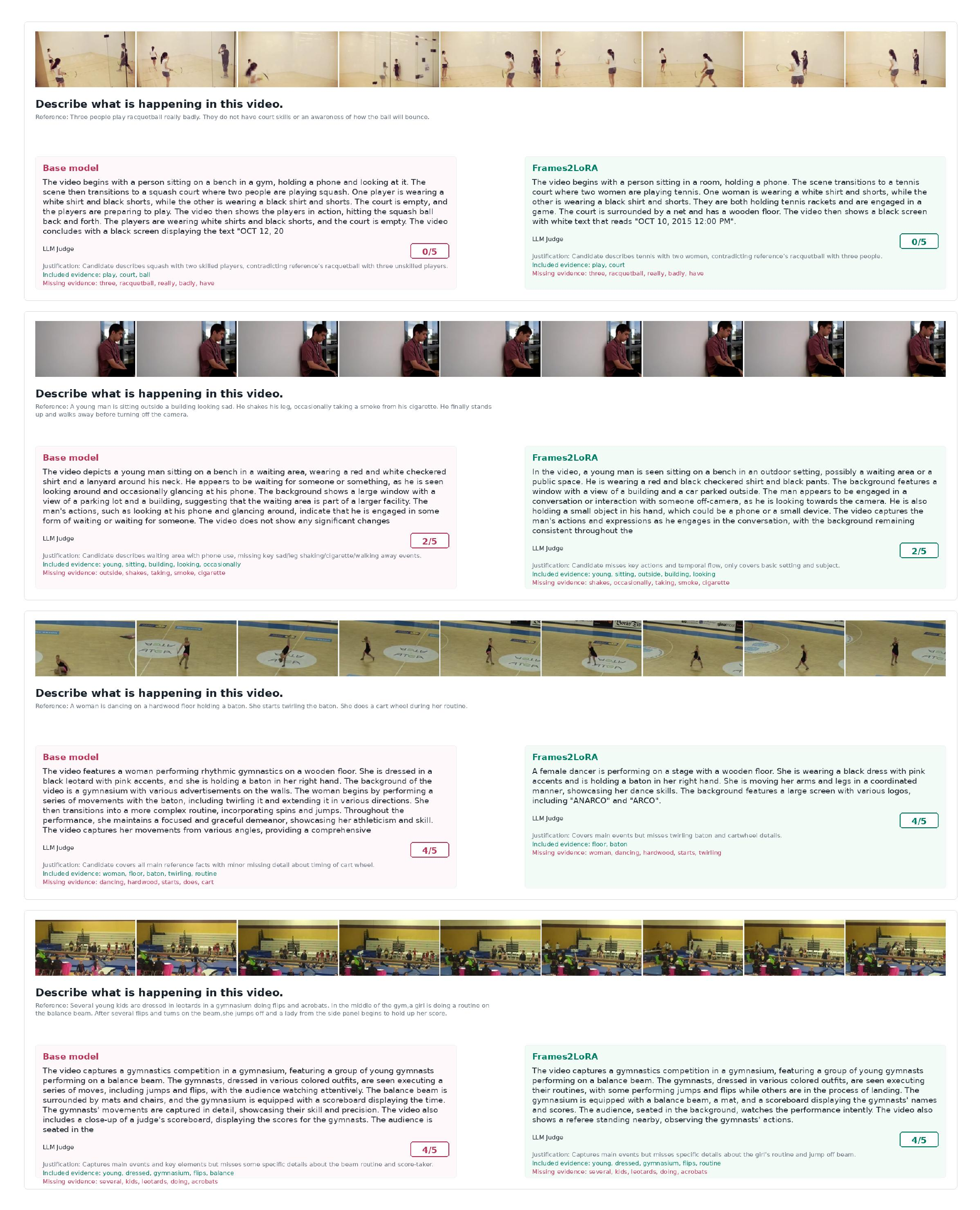}
  \caption{Qualitative examples from ActivityNet Captions.}
  \label{fig:qual_score_sweep:activitynet_captions}
\end{figure*}

\begin{figure*}[t]
  \centering
  \includegraphics[width=\linewidth]{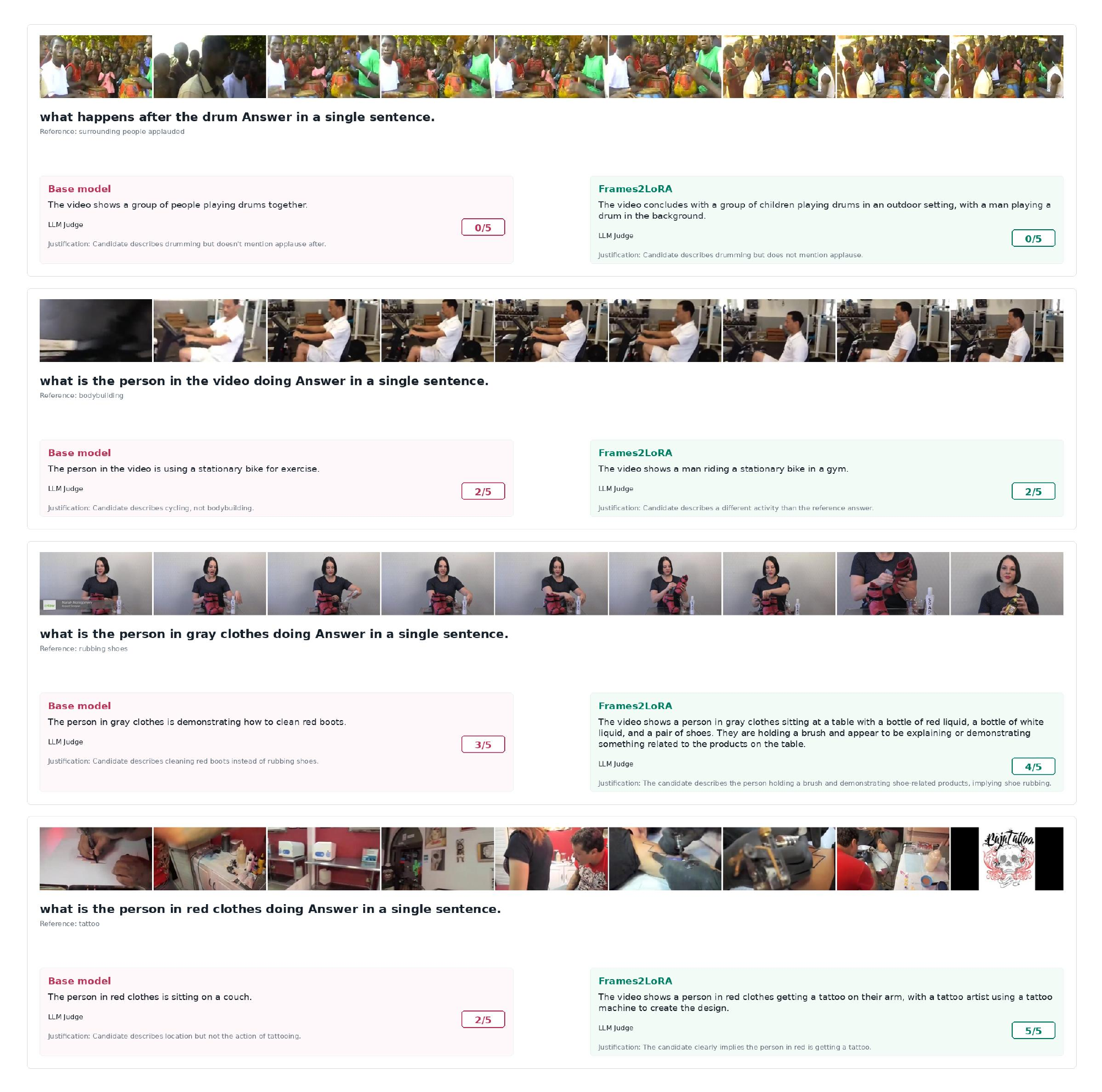}
  \caption{Qualitative examples from ActivityNetQA.}
  \label{fig:qual_score_sweep:activitynetqa}
\end{figure*}

\begin{figure*}[t]
  \centering
  \includegraphics[width=\linewidth]{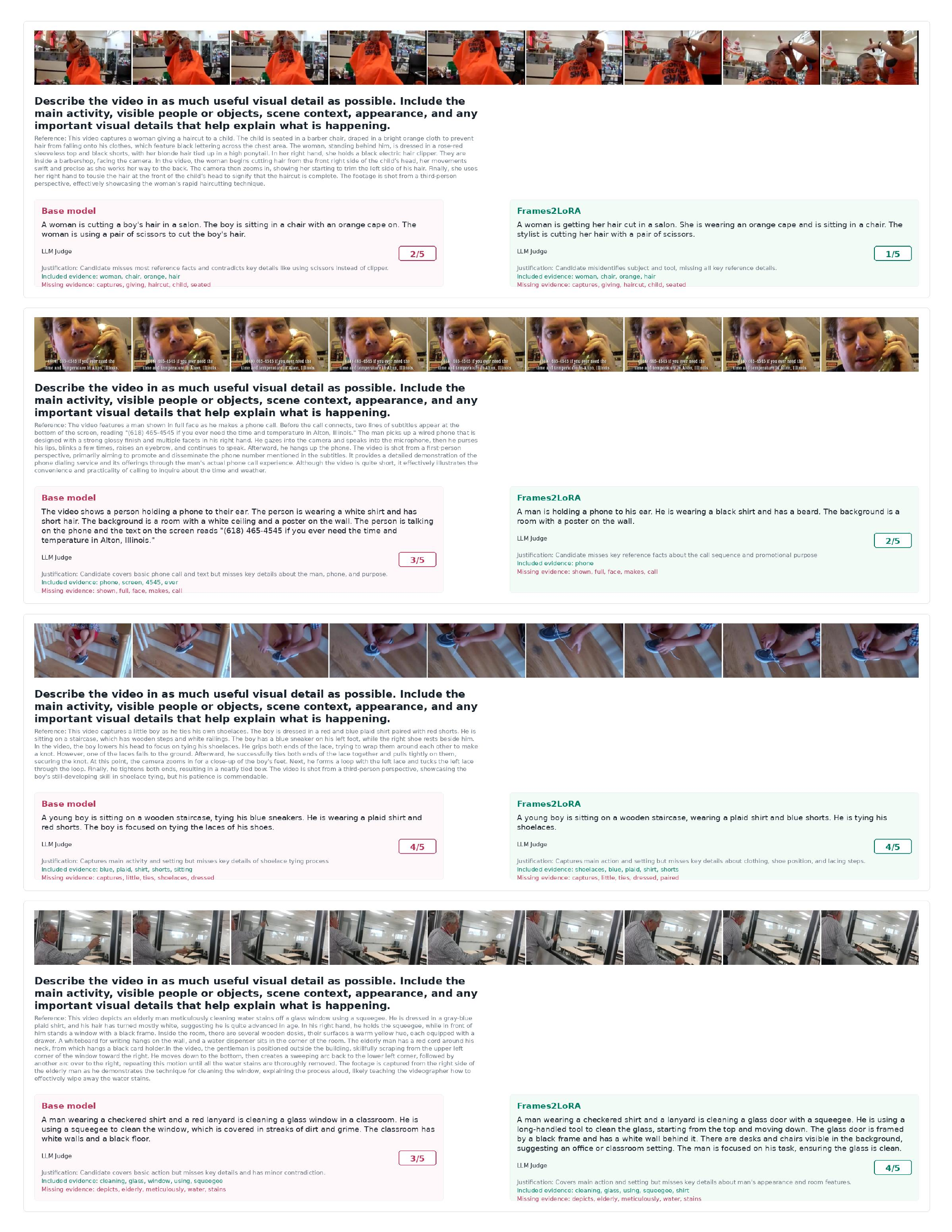}
  \caption{Qualitative examples from CaReBench: Caption.}
  \label{fig:qual_score_sweep:carebench_caption}
\end{figure*}

\begin{figure*}[t]
  \centering
  \includegraphics[width=\linewidth]{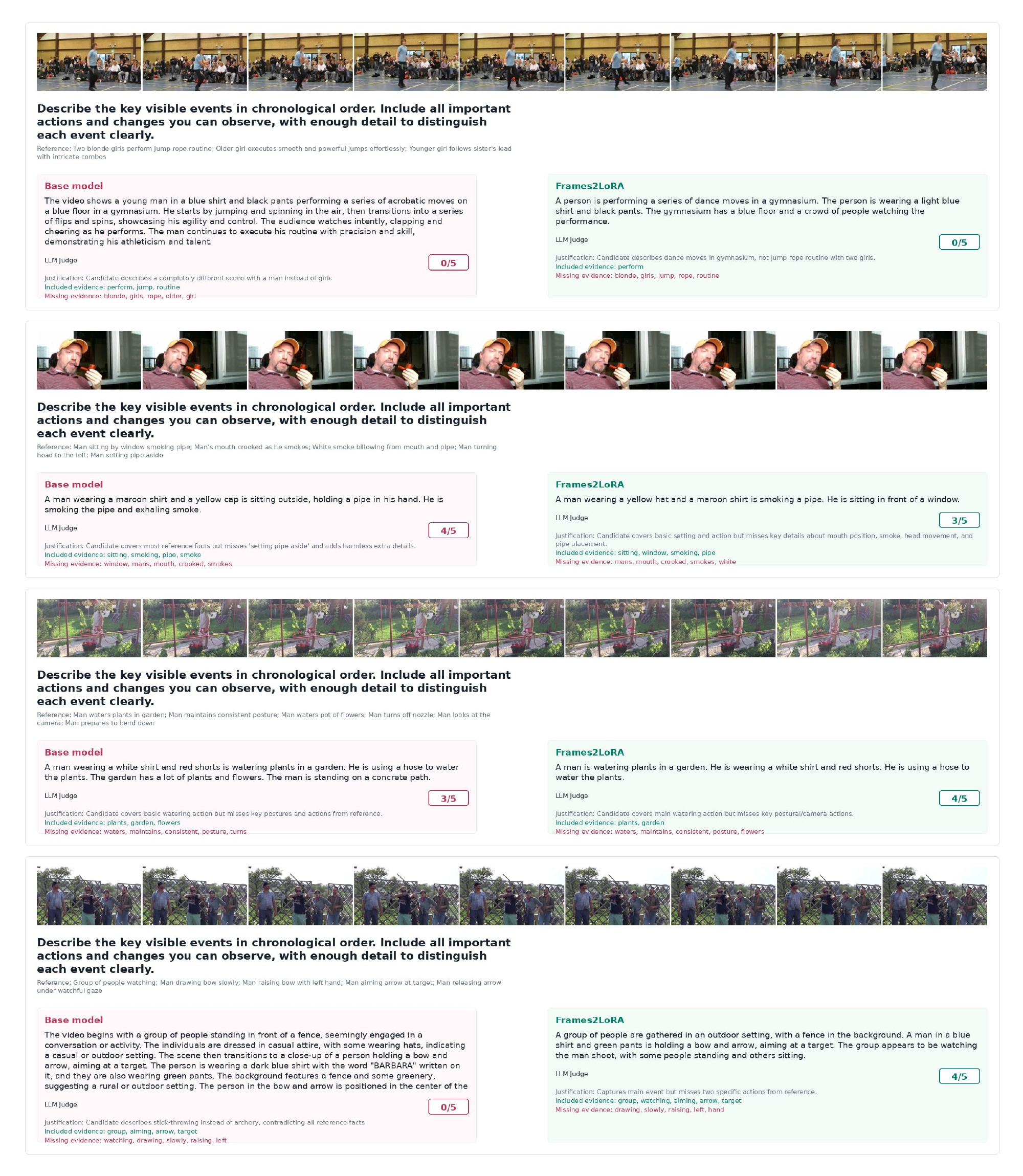}
  \caption{Qualitative examples from CaReBench: Events.}
  \label{fig:qual_score_sweep:carebench_events}
\end{figure*}

\begin{figure*}[t]
  \centering
  \includegraphics[width=\linewidth]{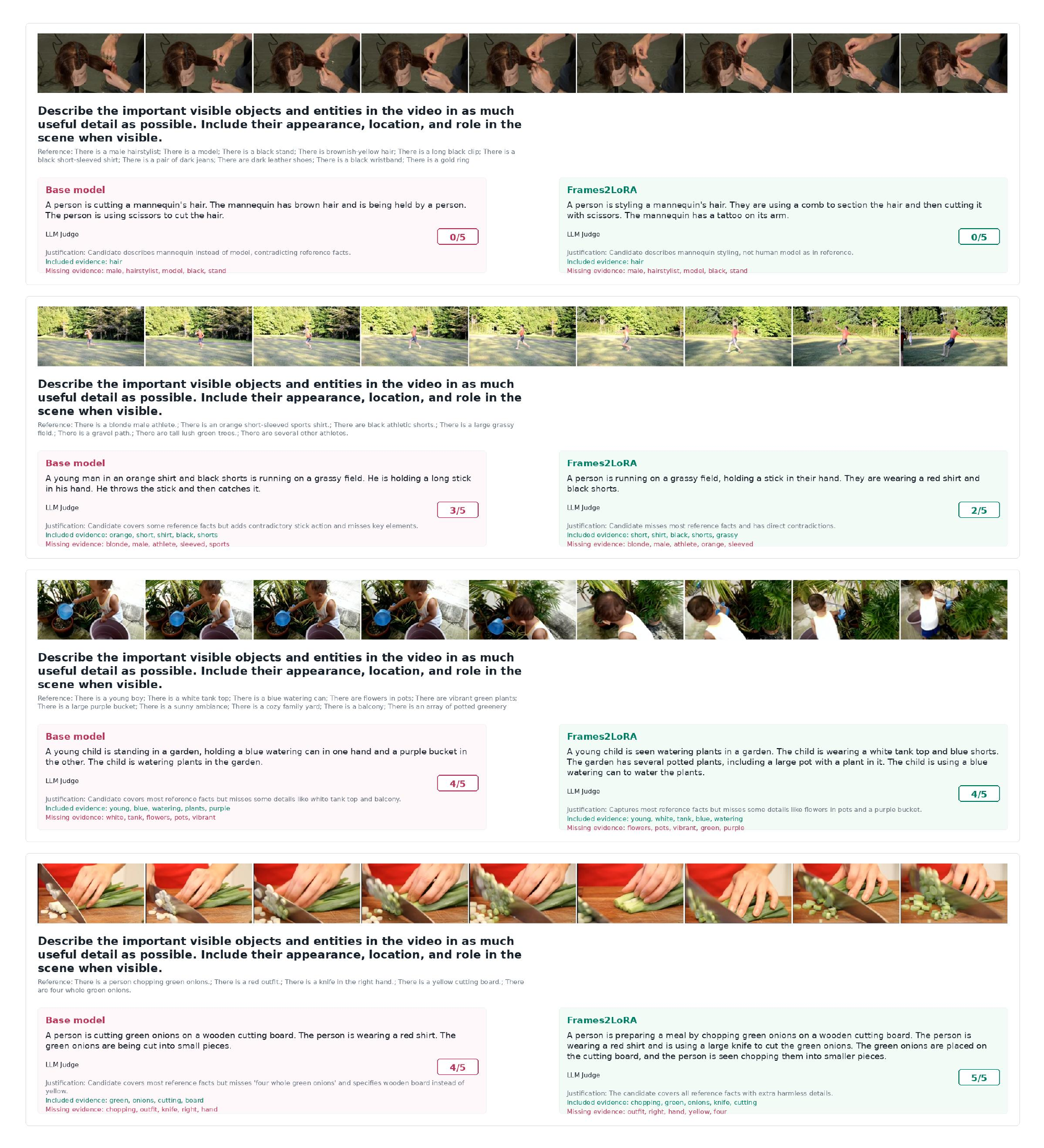}
  \caption{Qualitative examples from CaReBench: Objects.}
  \label{fig:qual_score_sweep:carebench_objects}
\end{figure*}

\begin{figure*}[t]
  \centering
  \includegraphics[width=\linewidth]{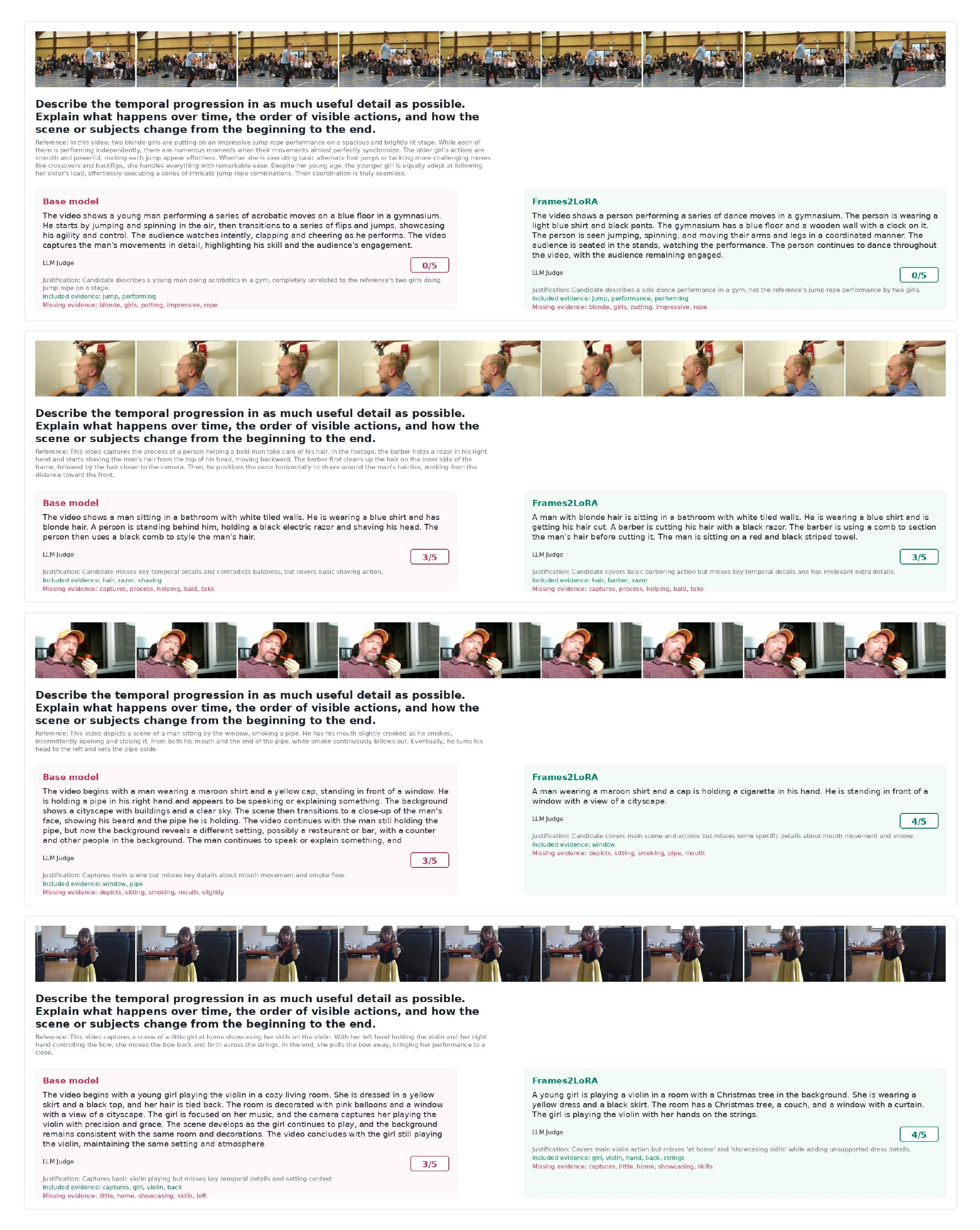}
  \caption{Qualitative examples from CaReBench: Temporal Caption.}
  \label{fig:qual_score_sweep:carebench_temporal_caption}
\end{figure*}

\begin{figure*}[t]
  \centering
  \includegraphics[width=\linewidth]{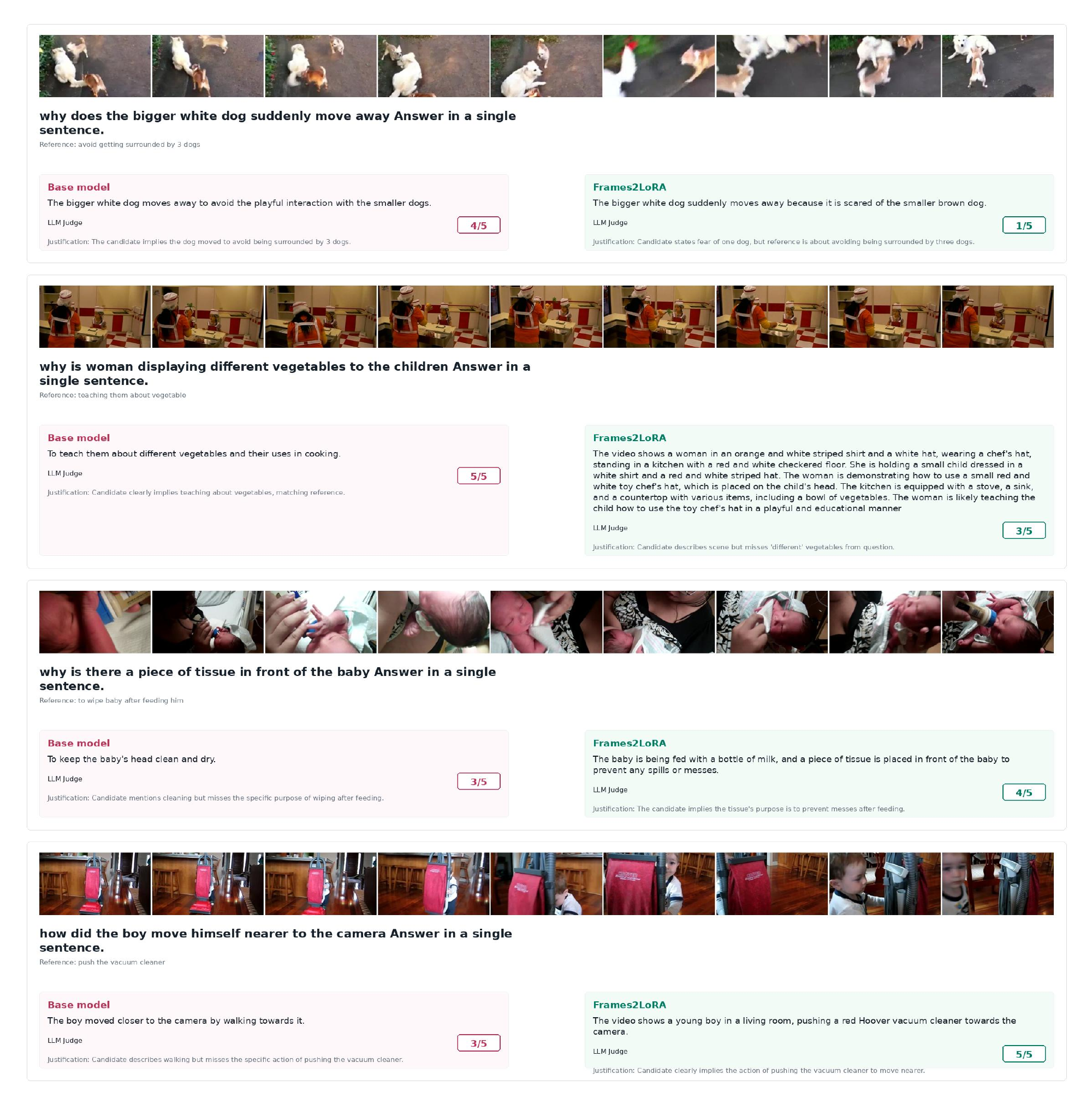}
  \caption{Qualitative examples from NExT-QA.}
  \label{fig:qual_score_sweep:next_qa}
\end{figure*}

\begin{figure*}[t]
  \centering
  \includegraphics[width=\linewidth]{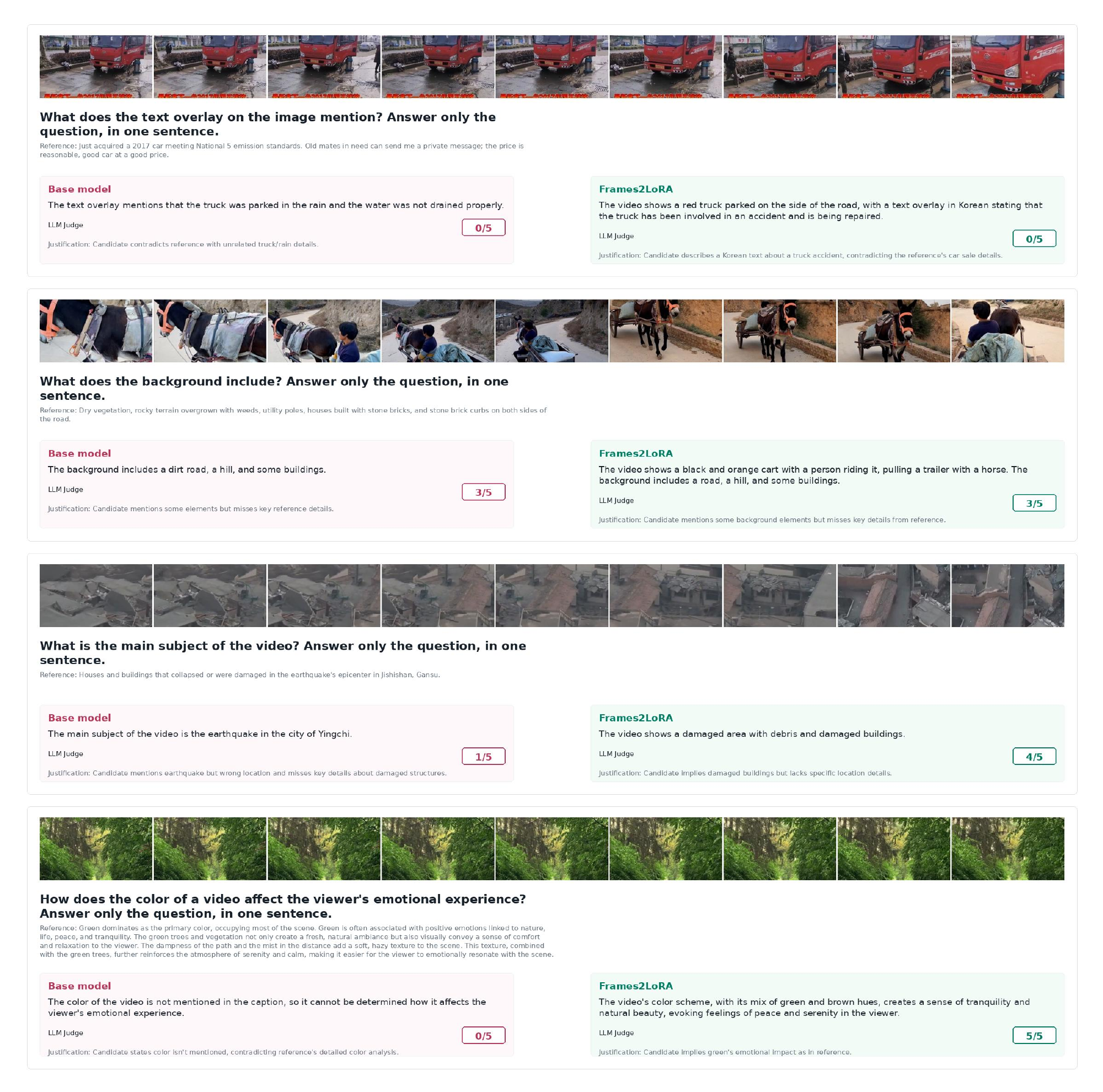}
  \caption{Qualitative examples from VidCapBench.}
  \label{fig:qual_score_sweep:vidcapbench}
\end{figure*}

\begin{figure*}[t]
  \centering
  \includegraphics[width=\linewidth]{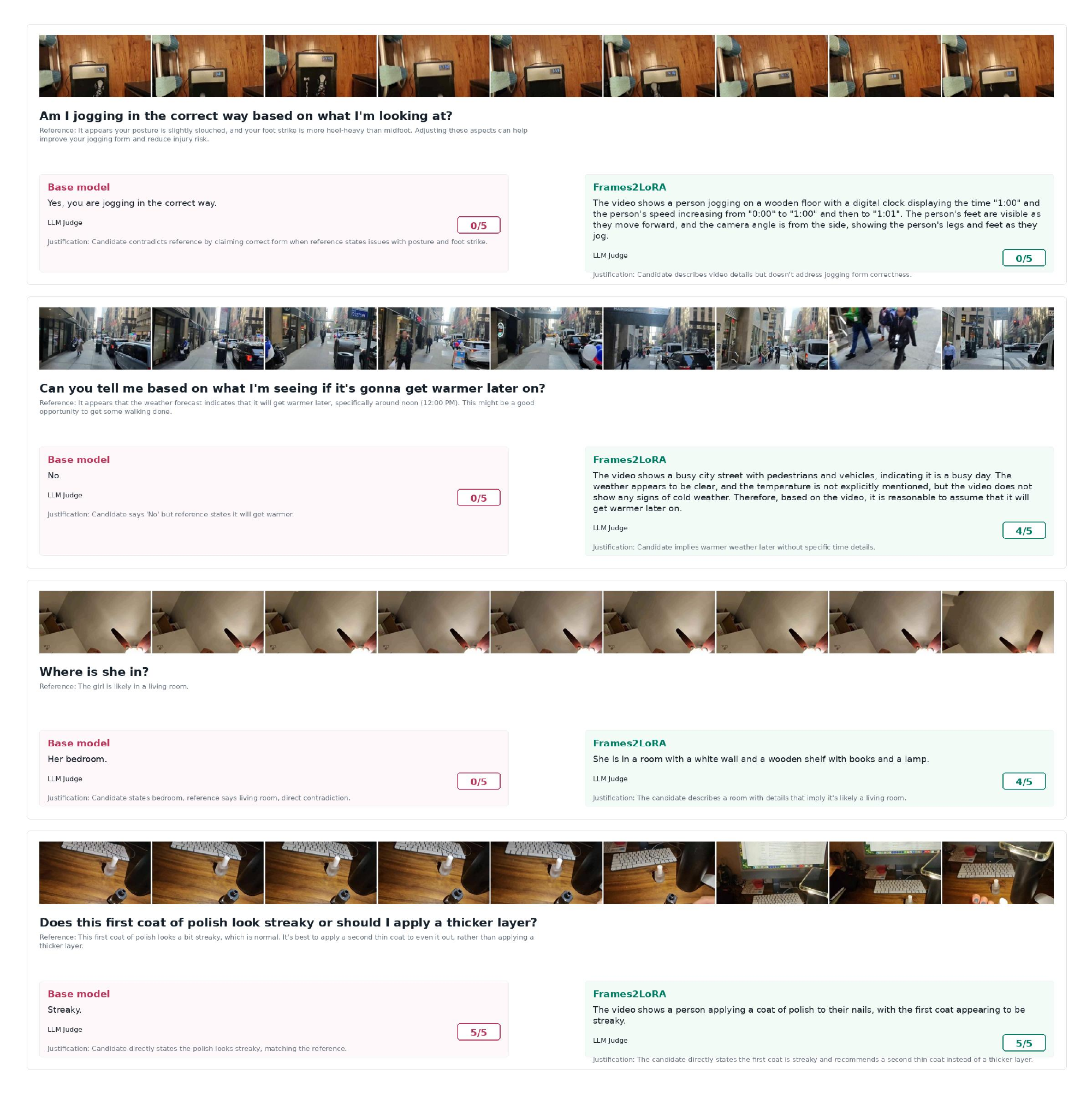}
  \caption{Qualitative examples from PLM SGQA.}
  \label{fig:qual_score_sweep:plm_sgqa}
\end{figure*}

\begin{figure*}[t]
  \centering
  \includegraphics[width=\linewidth]{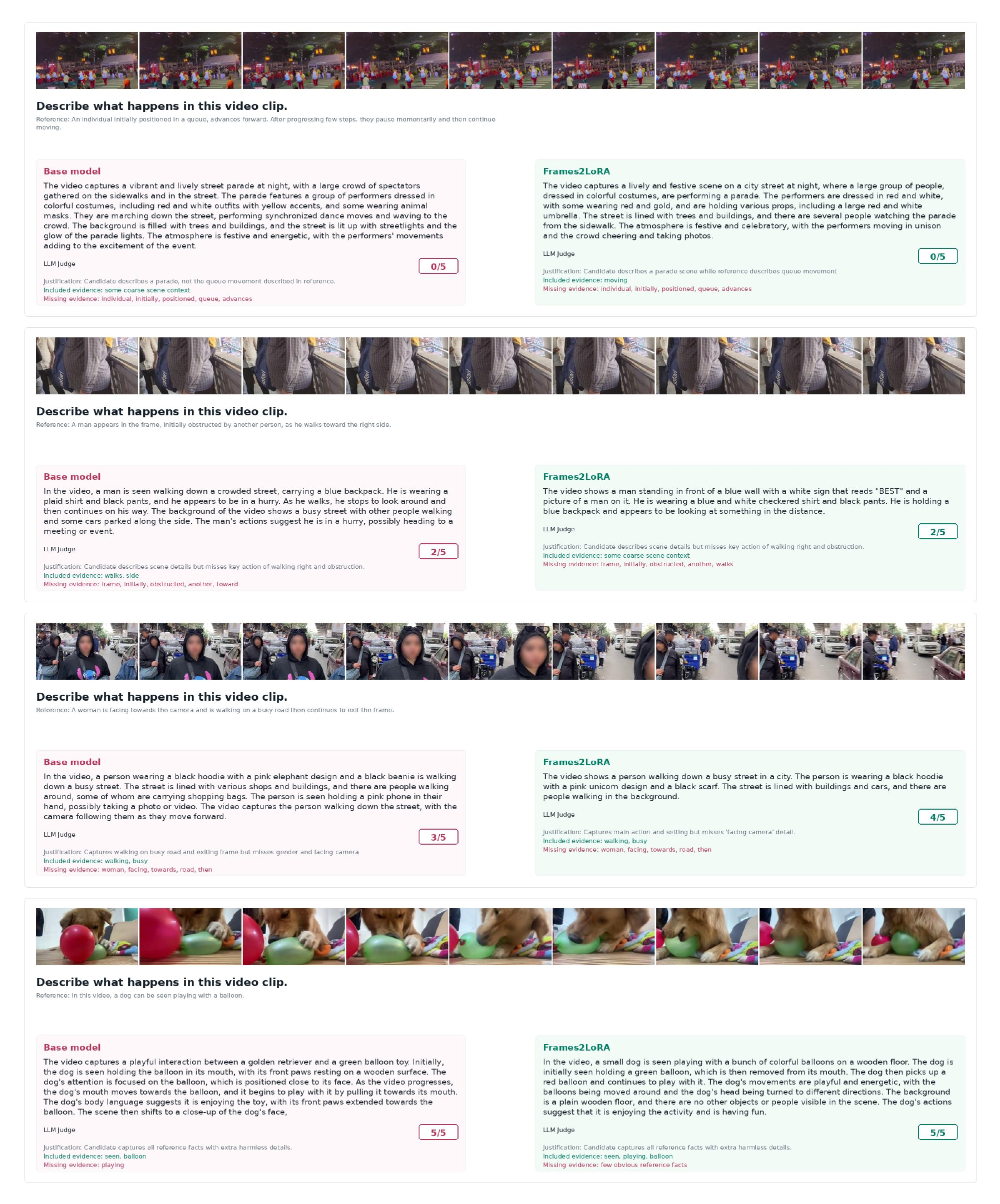}
  \caption{Qualitative examples from RCAP.}
  \label{fig:qual_score_sweep:rcap}
\end{figure*}

\begin{figure*}[t]
  \centering
  \includegraphics[width=\linewidth]{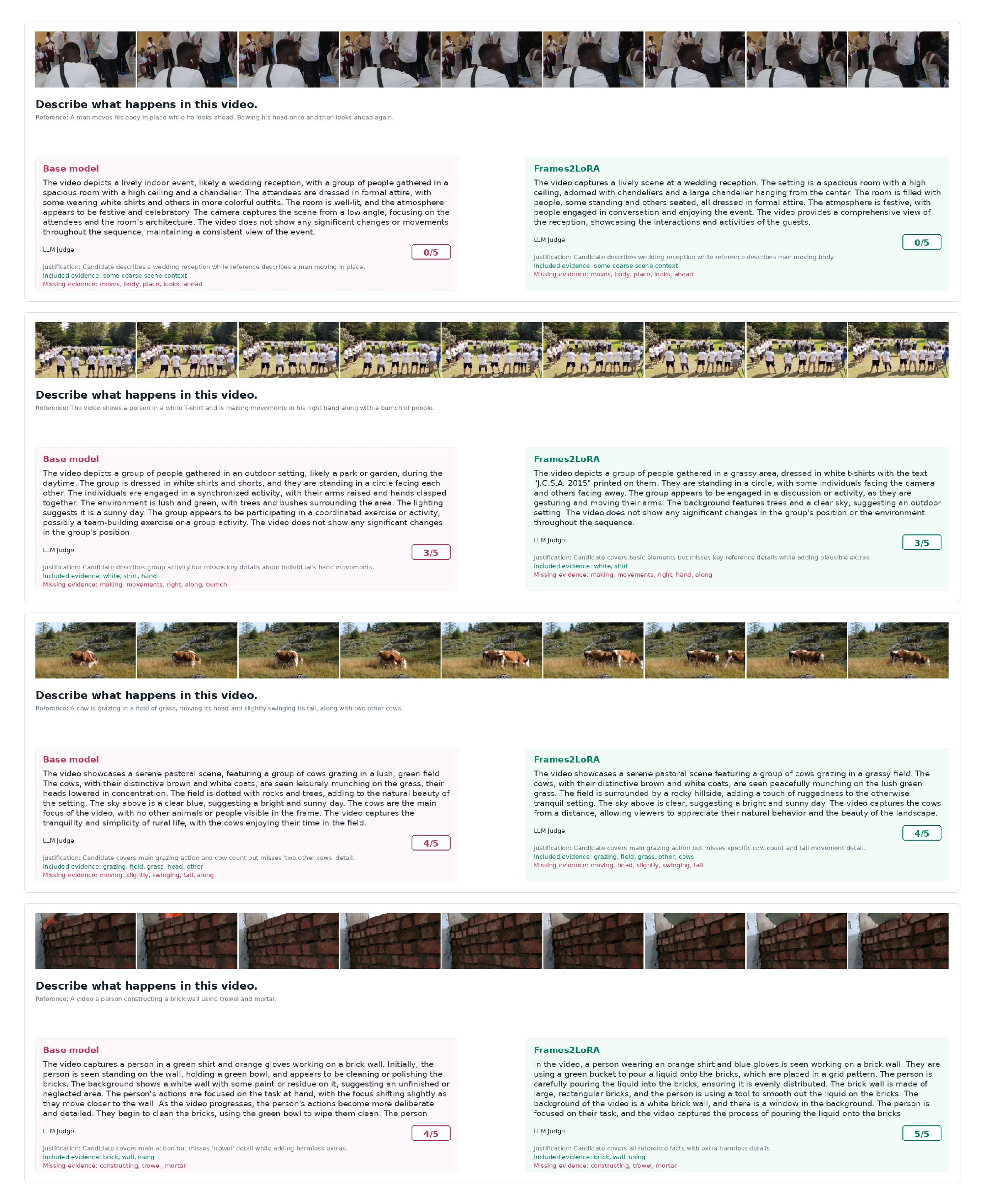}
  \caption{Qualitative examples from RDCAP.}
  \label{fig:qual_score_sweep:rdcap}
\end{figure*}

\begin{figure*}[t]
  \centering
  \includegraphics[width=\linewidth]{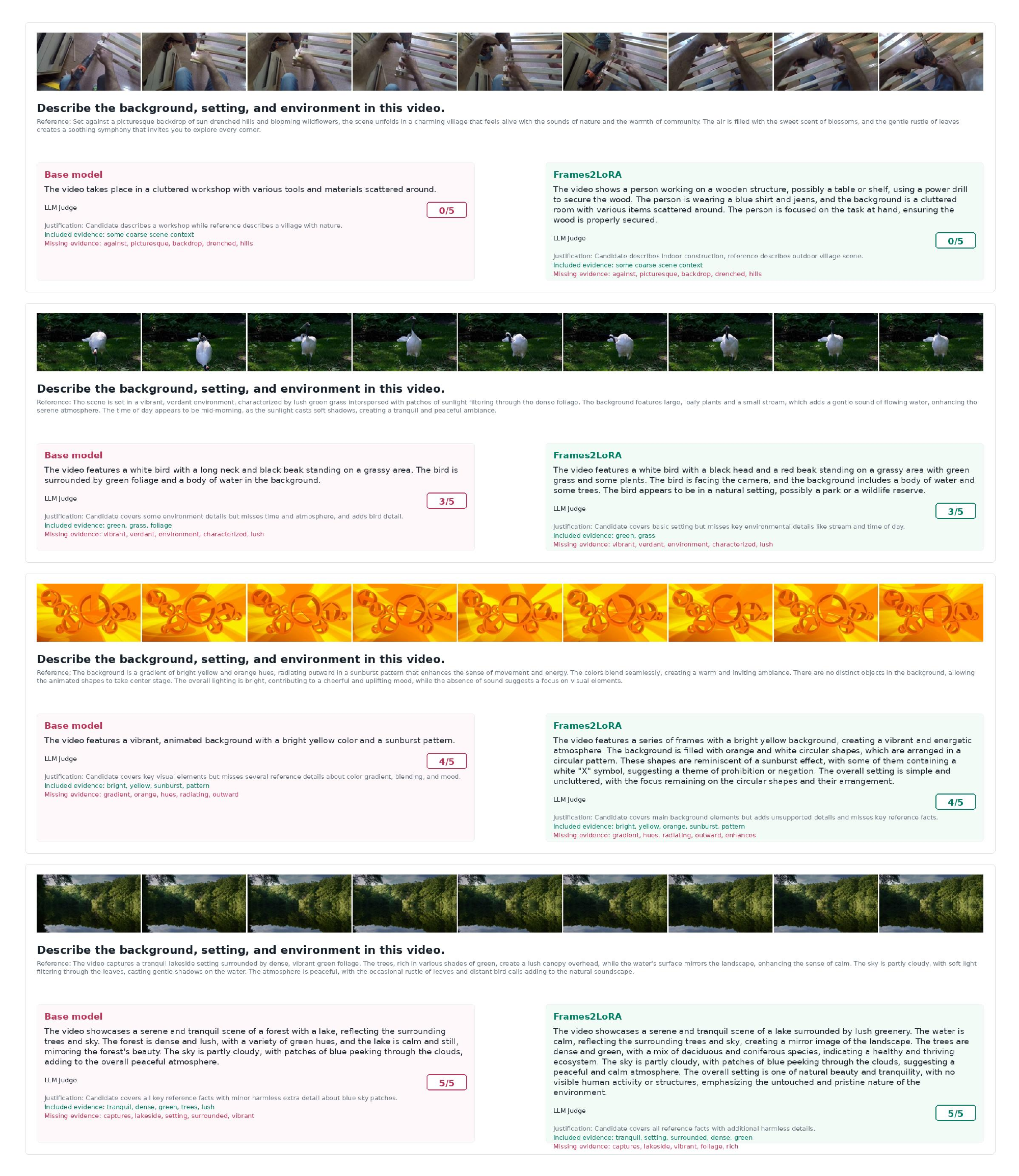}
  \caption{Qualitative examples from VDC Background.}
  \label{fig:qual_score_sweep:vdc_background}
\end{figure*}

\begin{figure*}[t]
  \centering
  \includegraphics[width=\linewidth]{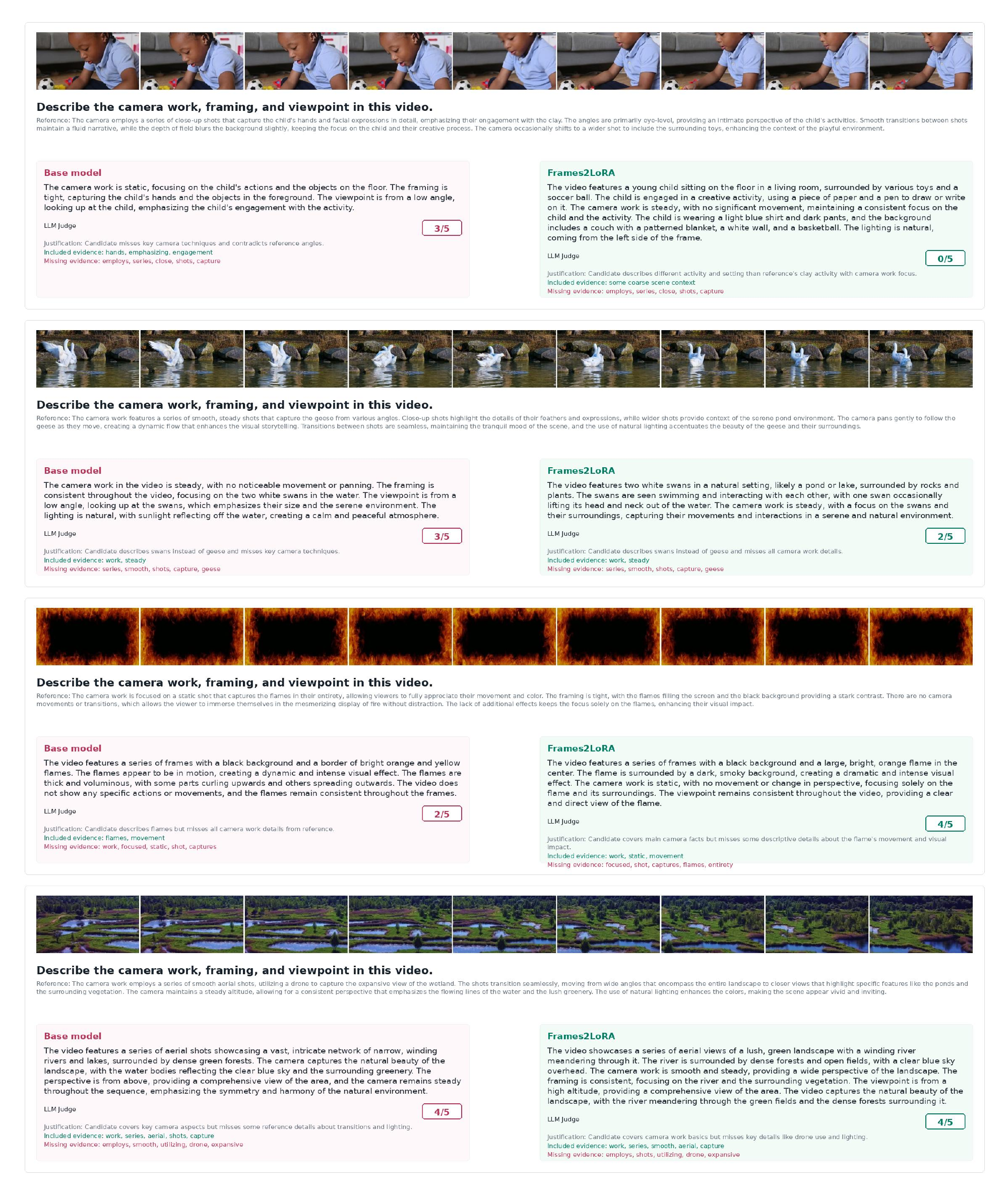}
  \caption{Qualitative examples from VDC Camera.}
  \label{fig:qual_score_sweep:vdc_camera}
\end{figure*}

\begin{figure*}[t]
  \centering
  \includegraphics[width=\linewidth]{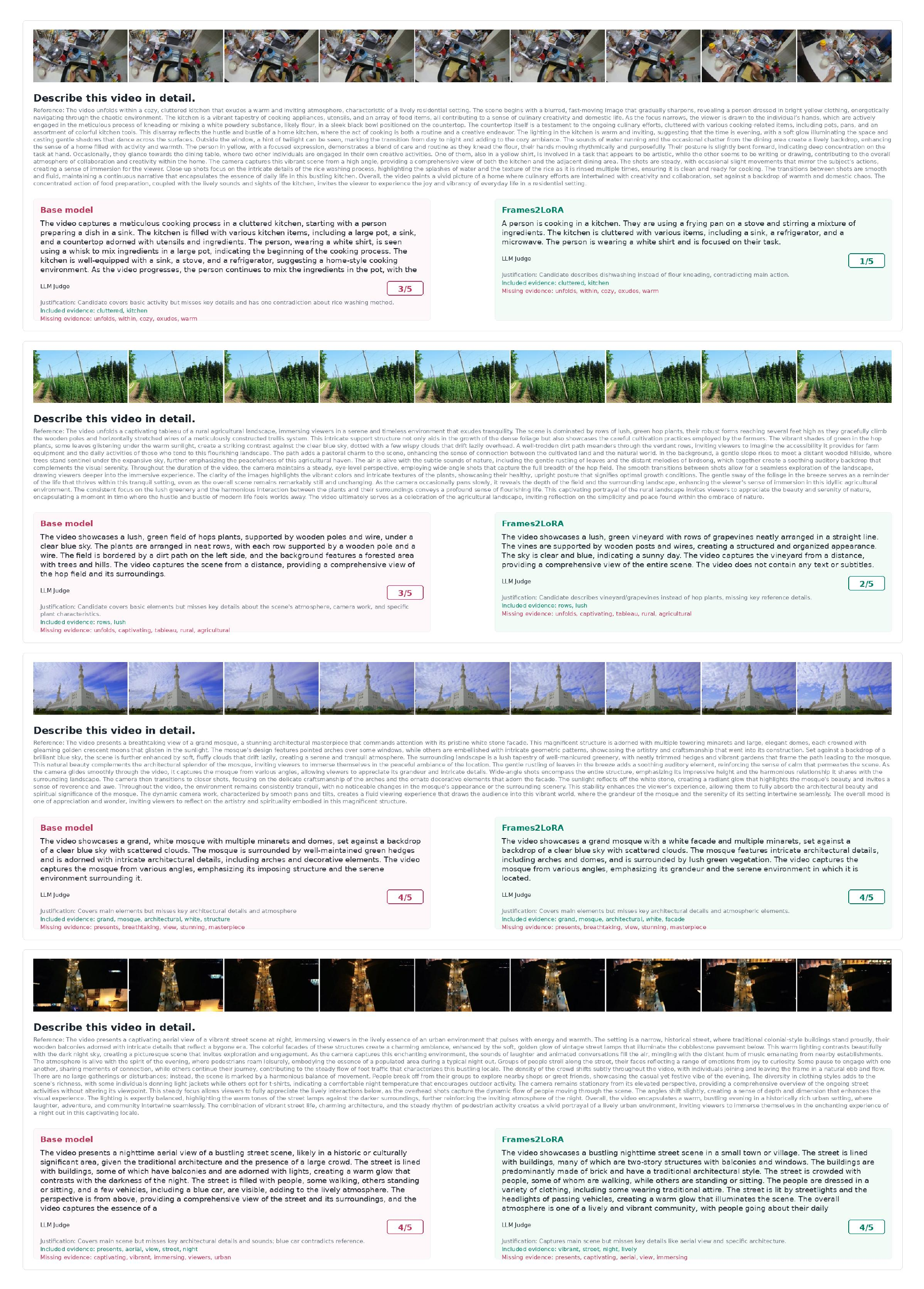}
  \caption{Qualitative examples from VDC Detailed.}
  \label{fig:qual_score_sweep:vdc_detailed}
\end{figure*}

\begin{figure*}[t]
  \centering
  \includegraphics[width=\linewidth]{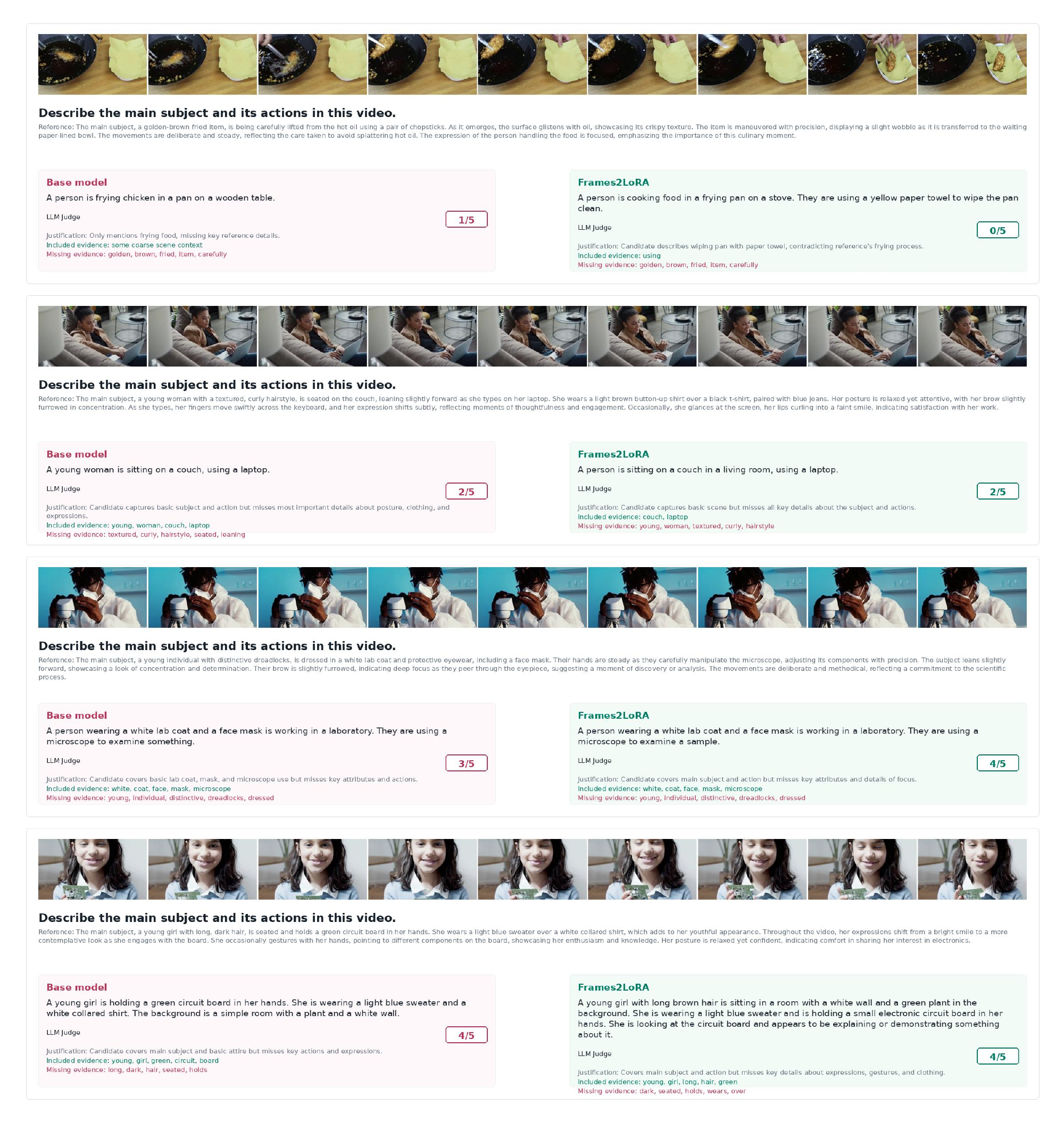}
  \caption{Qualitative examples from VDC Main Object.}
  \label{fig:qual_score_sweep:vdc_main_object}
\end{figure*}

\begin{figure*}[t]
  \centering
  \includegraphics[width=\linewidth]{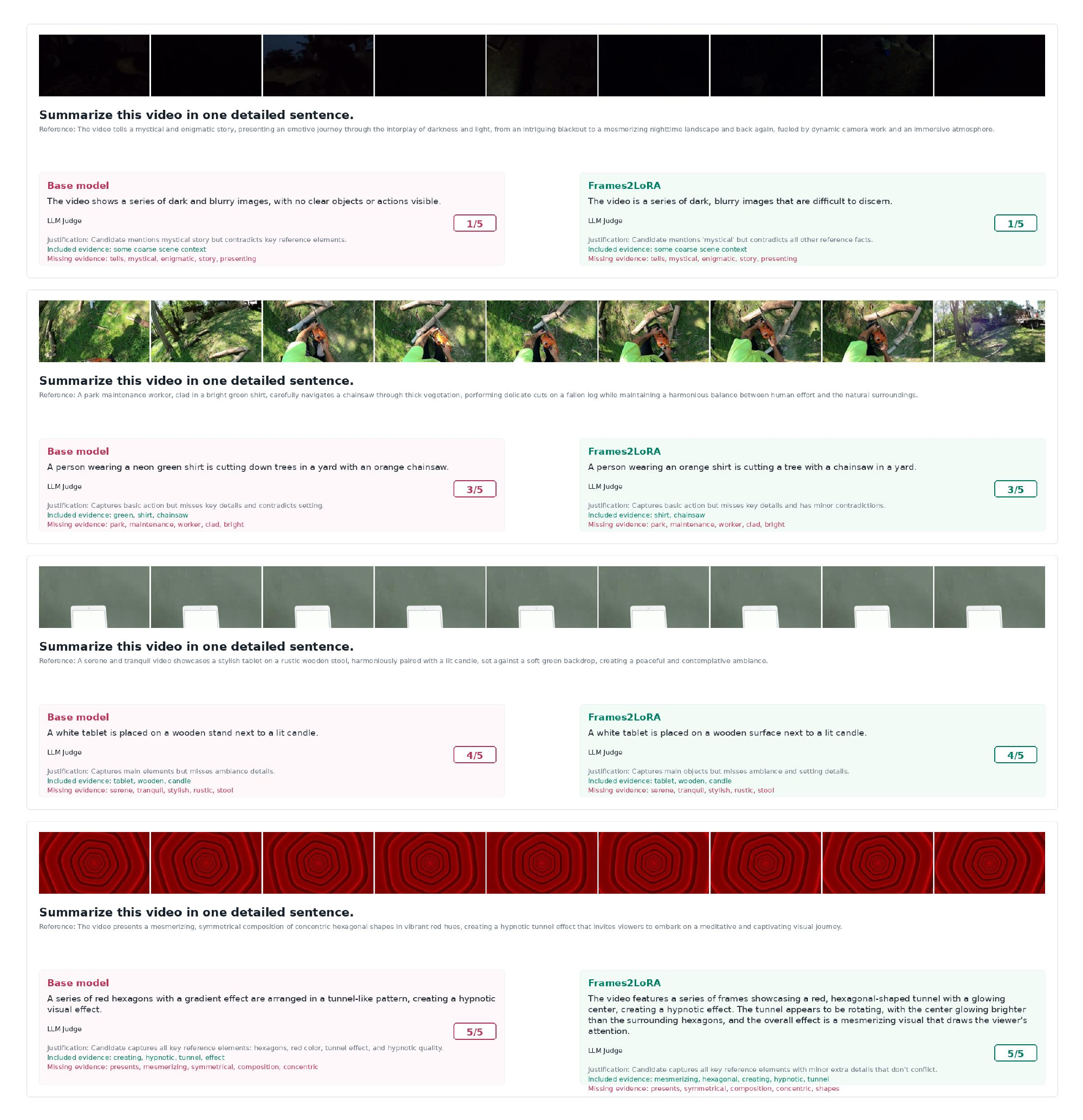}
  \caption{Qualitative examples from VDC Short.}
  \label{fig:qual_score_sweep:vdc_short}
\end{figure*}

Qualitative examples are shown in Figure~\ref{fig:qual_score_sweep:activitynet_captions}, Figure~\ref{fig:qual_score_sweep:activitynetqa}, Figure~\ref{fig:qual_score_sweep:carebench_caption}, Figure~\ref{fig:qual_score_sweep:carebench_events}, Figure~\ref{fig:qual_score_sweep:carebench_objects}, Figure~\ref{fig:qual_score_sweep:carebench_temporal_caption}, Figure~\ref{fig:qual_score_sweep:next_qa}, Figure~\ref{fig:qual_score_sweep:plm_sgqa}, Figure~\ref{fig:qual_score_sweep:rcap}, Figure~\ref{fig:qual_score_sweep:rdcap}, Figure~\ref{fig:qual_score_sweep:vdc_background}, Figure~\ref{fig:qual_score_sweep:vdc_camera}, Figure~\ref{fig:qual_score_sweep:vdc_detailed}, Figure~\ref{fig:qual_score_sweep:vdc_main_object}, Figure~\ref{fig:qual_score_sweep:vdc_short}, Figure~\ref{fig:qual_score_sweep:vidcapbench}.

% \begin{figure*}[t]
%   \centering
%   \includegraphics[width=\linewidth]{figures/A_qualitative_examples/carebench_spatial_caption__lora_score_spectrum.pdf}
%   \caption{Qualitative examples from CaReBench: Spatial Caption.}
%   \label{fig:qual_score_sweep:carebench_spatial_caption}
% \end{figure*}

\end{document}